\documentclass[letterpaper]{article} 
\usepackage{aaai24}  
\usepackage{times}  
\usepackage{helvet}  
\usepackage{courier}  
\usepackage[hyphens]{url}  
\usepackage{graphicx} 
\usepackage{natbib}  
\usepackage{caption} 
\usepackage{algorithm}
\usepackage{algorithmic}
\usepackage{xspace}
\usepackage{soul} 

\usepackage[dvipsnames]{xcolor}
\usepackage{tikz}
\usetikzlibrary{arrows.meta}
\usetikzlibrary{decorations.pathreplacing}

\usepackage{newfloat}
\usepackage{listings}
\DeclareCaptionStyle{ruled}{labelfont=normalfont,labelsep=colon,strut=off} 
\floatstyle{ruled}
\newfloat{listing}{tb}{lst}{}
\floatname{listing}{Listing}
\title{Introducing Delays in Multi-Agent Path Finding}
\author {
    Justin Kottinger\textsuperscript{\rm 1},
    Tzvika Geft\textsuperscript{\rm 2},
    Shaull Almagor\textsuperscript{\rm 3},
    Oren Salzman\textsuperscript{\rm 3},
    Morteza Lahijanian\textsuperscript{\rm 1}
}
\affiliations {
    \textsuperscript{\rm 1} Department of Aerospace Engineering, University of Colorado Boulder, USA\\
    \textsuperscript{\rm 2} The Blavatnik School of Computer Science, Tel Aviv University, Israel\\
    \textsuperscript{\rm 3} The Henry and Marilyn Taub Faulty of Computer Science, Technion, Israel\\
    justin.kottinger@colorado.edu, zvigreg@mail.tau.ac.il, shaull@technion.ac.il, osalzman@cs.technion.ac.il, morteza.lahijanian@colorado.edu
}

\usepackage{amsmath}
\usepackage{amsthm}
\usepackage{amssymb}
\usepackage{enumitem}
\usepackage{complexity}
\usepackage{todonotes}
\usepackage[capitalise]{cleveref}
\usepackage{caption}
\usepackage{subcaption}
\usepackage{graphicx}
\usepackage{multirow}
\usepackage{booktabs}
\usepackage{siunitx}
\newcommand{\bbN}{\mathbb{N}}

\newcommand{\tup}[1]{\langle #1 \rangle}
\newcommand{\length}{\boldsymbol{\ell}}

\newtheorem{theorem}{Theorem}
\newtheorem{remark}{Remark}
\newtheorem{problem}{Problem}
\newtheorem{lemma}{Lemma}
\newtheorem{example}{Example}

\newtheorem{observation}{Observation}

\newcommand{\mapf}{MAPF\xspace}

\begin{document}
\maketitle
\begin{abstract}
We consider a Multi-Agent Path Finding (\mapf) setting where agents have been assigned a plan, but during its execution some agents are delayed. Instead of replanning from scratch when such a delay occurs, we propose \emph{delay introduction}, whereby we delay some additional agents so that the remainder of the plan can be executed safely.
We show that finding the minimum number of additional delays is \APX-hard, i.e., it is \NP-hard to find a $(1+\varepsilon)$-approximation for some $\varepsilon>0$.
However, in practice we can find \emph{optimal} delay-introductions using Conflict-Based Search for very large numbers of agents, and both planning time and the resulting length of the plan are comparable, and sometimes outperform the state-of-the-art heuristics for replanning.
\end{abstract}

\section{Introduction}
\label{sec:intro}

Multi-Agent Path Finding (\mapf) is a
problem in Artificial Intelligence (AI) that asks to find non-colliding paths for a group of agents moving on a graph~\cite{stern2019multi,SalzmanS20}. 
Applications vary from autonomous warehouse management~\cite{wurman2008cooperative} and factory pipe routing~\cite{BelovDBHKW20} to rail planning~\cite{LCZCHS0K21} and swarm robotics \cite{7395308}.
Although \mapf is known to be generally an intractable problem~\cite{Yu2016,Banfi2017,nebel2020computational, DBLP:conf/socs/Geft23}, recent algorithms can scale to thousands of agents, e.g., 
\cite{Li_Chen_Harabor_Stuckey_Koenig_2022,Li_Ruml_Koenig_2021,Okumura_2023}.
A limiting aspect of these algorithms is the simplifying assumption that, at deployment, agents can synchronously execute a plan.  In reality, however, it is common for agents to fall out of sync, e.g., due to delays or model uncertainty.  Such incidents may cause the plan to no longer be valid (non-colliding), in which case we must either compute a new plan or repair the old plan quickly. This is challenging  since replanning faces the same difficulties as the original \mapf problem, and plan repair is shown to be as difficult as  plan generation itself \cite{nebel1995plan}.

In this work, we propose a simple but effective approach to plan repair that inherits a lot of the benefits of the original plan and can scale to a large number of agents.  
Our key idea is to use the topology of the original plan and resolve conflicts by allowing agents to stay in place.  That is, we repair the plan  by introducing \emph{delays} -- i.e., requiring some agents to remain in a certain location for a certain amount of time instead of advancing according to the prescribed path.
The intuition is that resources are put into generating and validating the original plan.  It is hence desirable to maintain at least some of its properties.  
For example, in safety-critical or ethical situations (e.g., transportation of hazardous materials, air traffic control) plans often need to be approved by a human controller, and therefore, replanning requires the human to accept a new plan, which in turn requires trust. 
By using the same paths, we gain several benefits, including inheriting the visual explainability of the original plan (i.e., the paths visually remain the same), reducing the search space to a smaller graph than the original one, and existence of a solution when delays are not constrained along the paths.


Specifically, we consider the following setting: we work over an environment modeled as a directed graph, where agents wish to move from their starting vertices to their goals. We further assume that we already have a plan $P$ that drives each agent from start to goal. However, $P$ may contain collisions. Motivationally, we think of $P$ as obtained from a non-colliding plan by having some agents delay in place, resulting in possible collisions. We allow to repair $P$ to a new plan $P'$ by having some agents delay at certain vertices.
Furthermore, we want $P'$ to be such that the overall number of delays is minimal.

We also draw motivation from a non-optimization problem formulation closely related to ours by \cite{abrahamsen2023coordination}.
They focus on adding \emph{any} set of delays to $P$ to ensure $P'$ is non-colliding, for which they provide positive and negative complexity results. Notably, they mention studying optimization variants, akin to our problem, as an open extension.

Introducing a minimal set of delays in this setting gives rise to some intricate behaviors, as demonstrated in the following examples. In particular, the choice of when to delay an agent and for how many steps is crucial and nontrivial.

\begin{figure}
    \centering
    \begin{subfigure}{0.49\linewidth}
        \centering    \includegraphics[width=\columnwidth]{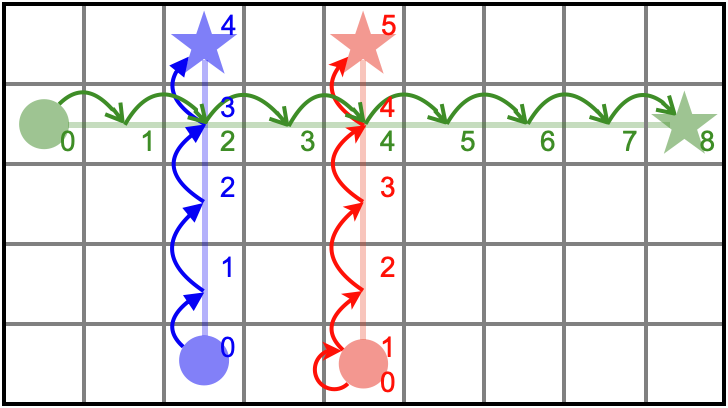}
        \caption{\cref{xmp:postpone}}
        \label{fig:example_delay_late}
    \end{subfigure}
    \hfill
    \begin{subfigure}{0.49\linewidth}
        \centering
        \includegraphics[width=\columnwidth]{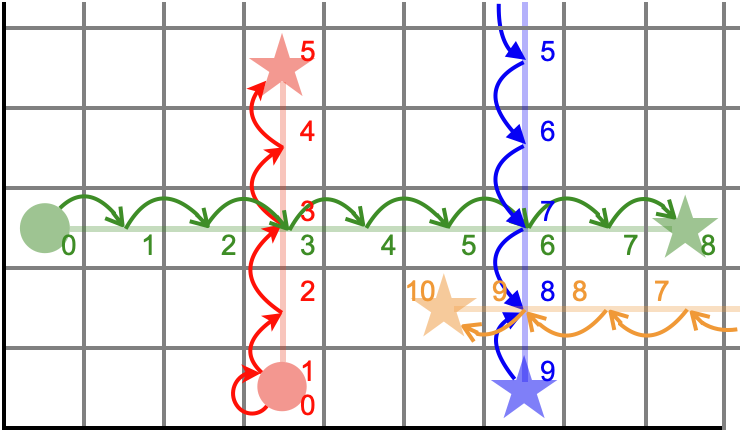}
        \caption{\cref{xmp:multiple_delays}}
        \label{fig:example_delay_twice}
    \end{subfigure}
    \hfill
    \caption{Setting of (a) \cref{xmp:postpone} and (b) \cref{xmp:multiple_delays}. Straight lines depict the original
plan, and curved edges represent the plan when Red agent is delayed at time 0. (a) It is better to have Green agent delay at time 3. (b) it is better to have Green agent delay for
two timesteps.}
    \label{fig:fun_examples}
\end{figure}

\begin{example}[Postponing delays]
\label{xmp:postpone}
Consider the case in Fig.~\ref{fig:example_delay_late}, which shows a plan for three agents.  Note that the original plan, which takes the agents straight to their respective goals, is non-colliding.  Now, imagine 
Red is delayed at time 0, resulting in the detection of an upcoming collision with Green at time 4. However, delaying Green upon detection (step 1) will cause a collision with Blue at time 3. Instead, it is preferable to postpone the delay of Green to time~3, and let Blue pass through.
\end{example}

\begin{example}[Long delays]
\label{xmp:multiple_delays}
Consider the setting in \cref{fig:example_delay_twice}. Green is again about to collide with Red. However, delaying Green for a single timestep will cause another collision with Blue at time 7. Blue, in turn, passes a train of 100 agents (depicted in orange) just before it crosses at time 8. Thus, if we delay Blue for even a single timestep, this would require either delaying it for another 100 times, or delaying the train of 100 agents. Thus, the optimal solution is to delay Green for 2 timesteps or delay Red for another timestep.
\end{example}
\noindent 
These examples allude to our two main contributions: 
\begin{itemize}
    \item We show that the problem of \emph{avoiding collisions by introducing delays} (ACID) is \NP-complete,
    and in fact even \APX-hard.

    \item We propose an algorithmic approach to ACID by formulating it as a small instance of \mapf, with a low branching degree, for which existing algorithms can be readily used. 
    
\end{itemize}
We experimentally evaluate our algorithmic approach through several standard benchmark problems. Specifically, the results show that for a small number of unexpected delays (one delay, in our experiments), the simplicity of the small \mapf problem allows to use optimal algorithms such as Conflict-based Search (CBS) \cite{sharon2015conflict} to compute a plan with minimum number of delays fast and scale up to \num[group-separator={,}]{1000} agents. For more delays (with a large number of agents), CBS does not scale, but Heuristic-based \mapf algorithms perform well. 
\paragraph*{Related Work:}
The execution of MAPF plans may present unexpected delays that hinder the system's ability to follow the prescribed plan.
~\citet{AtzmonSFWBZ20} proposed a method to account for these uncertainties by computing $k$-robust plans (for some user-provided $k$) that guarantee safe execution even in the presence of up to $k$ delays. \citet{AtzmonSFSK20} also extended the idea to the probabilistic setting, guaranteeing success with probability at least $p$, when given a (user defined) probabilistic model of delays. However, these robust planning techniques 
suffer from being computationally expensive, overly-conservative, 
and are robust only in expectation (probabilistically).

Plan repair was considered almost three decades ago by~\citet{nebel1995plan}, where the authors show that repairing plans is potentially harder than planning from scratch. Interestingly, they show that a bottleneck of repairing plans is choosing the plan that we repair \emph{to}. 
This challenge can be avoided in specific cases. For example,~\citet{10013661} repaires single-agent paths in the presence of dynamic obstacles by connecting pre-computed path-segments together to repair an invalid motion plan. Their work does not consider coordination with additional agents. 
\citet{komenda2011multi} introduce the generalized problem of multi-agent plan repair and proposed three sub-optimal algorithms. \citet{KOMENDA201476} present an optimal way to solve the problem but their results only consider up to $10$ agents. 

\citet{Hoenig_Kumar_Cohen_Ma_Xu_Ayanian_Koenig_2016} solves problems associated with delays via a post-processing technique that transforms a MAPF plan into a plan-execution schedule (MAPF-POST). Their setting crucially relies on agents’ kinematic abilities. Specifically, the ability to use rational constant speeds allows reducing the problem to a linear program (LP) using simple temporal networks. That work is extended by \citet{berndt2020feedback} by formalizing the problem as a MILP and solving it sub-optimally online. Similarly, \citet{Ma_Kumar_Koenig_2017} present a probabilistic approach to resolving delays in decentralized systems by employing an approximate expectation-minimization approach. Note that casting our work as an instance of MAPF-POST allows for a simple solution: when an agent is delayed, simply slow down agents that may collide with it. This is a particularly simple case of MAPF-POST. Therefore, our main interest is in the combinatorial aspect of this problem in the discrete case.

There are recent works that are similar to our work. \citet{bartak2018scheduling} solve traditional MAPF via a scheduling-based approach. Specifically, they use a layered graph to represent delays. If only a single layer is used, no delays are allowed. Secondly, \citet{vsvancara2023multi} present a preliminary work where the goal is to solve MAPF by only introducing delays onto predefined paths. They create an abstract graph where the nodes represent agents and the edges represent choices to wait. Lastly, \citet{abrahamsen2023coordination} studies the most similar version of our problem.
There, however, 
they study the \emph{feasibility} problem of finding a set of delays that allows agents to reach their targets without colliding along a set of simple paths.
They provide an in-depth computational complexity investigation in lieu of empirical results.
Specifically, they present a sharp tractability boundary based on a key parameter called \emph{vertex multiplicity (VM)},
defined as the maximum number of paths passing through the same vertex. They present a variant of the problem that is NP-complete for $\textnormal{VM}=3$ and efficiently solvable for $\textnormal{VM} \le2$.

This work differs from all of these works in multiple aspects. First, in contrast to \citet{Hoenig_Kumar_Cohen_Ma_Xu_Ayanian_Koenig_2016} and \citet{berndt2020feedback}, we consider a combinatorial problem, which does not admit a reduction to LP. Moreover, our approach allows for constraints that prohibit certain delays. Secondly, we deviate from \citet{Ma_Kumar_Koenig_2017} by (i) considering a worst-case scenario rather than a probabilistic one and (ii) provide an optimal approach for the setting where the system has a centralized controller rather than proposing execution policies for decentralized systems. The key differences between the work of \citet{vsvancara2023multi} and our work are that (i) we use traditional MAPF algorithms rather than optimization techniques and (ii) our graph is comparatively very small, allowing us to solve much more complex problem instances in a much shorter amount of time. 

Additionally, our problem setting is very similar to that of \citet{vsvancara2023multi} but our solving techniques are quite different. Specifically, we avoid the feasibility problem by assuming the predefined paths came from an originally safe plan and solve the problem optimally using out-of-the-box MAPF solvers. And finally, in contrast to \citet{abrahamsen2023coordination}, we study the optimality of plans, i.e., we aim to minimize the number of delays that are introduced, and not merely check for the feasibility of some number of delay introductions. Indeed, in practical settings (as well as in our experimental setting), collisions occur due to agents breaking down. In such cases, it is trivial to repair the plan by introducing delays in all the remaining agents, thus halting execution until the breakdown is fixed (c.f., \cref{rmk:upper_bound_delays}).

In contrast to~\citet{AtzmonSFSK20,AtzmonSFWBZ20}, our method does not pre-compute a robust plan (and hence has no inherent added computational cost), but rather fixes (repairs) an existing plan if a delay occurs. We show theoretically that our setting still incurs the computational hardness presented by~\citet{nebel1995plan} (c.f., \cref{thm:ACID_NP_complete}). However, we mitigate practical computation by keeping the set of repaired plans relatively small due to strictly limiting the agents to only using delays. Thus, modern MAPF algorithms allow us to practically repair plans.

\section{Problem Statement}
\label{sec:problem}
We start by formulating the general MAPF setting.
Consider $n\in \bbN$ agents, acting in an environment represented by a directed graph $G=\tup{V, E}$ where each agent $i\in \{1,\ldots,n\}$ has a source $s_i \in V$ and a goal $g_i \in V$. 
A \emph{path} in $G$ is a sequence of vertices $\pi=v_1v_2\ldots v_m$  such that $(v_k, v_{k+1})\in E$ for all $1 \leq k < m$. 
We assume that the vertices of~$G$ contain self loops (i.e., for all $v\in V$, we have $(v,v) \in E$), so agents can be delayed. We remark that our results still hold if one allows self loops only on some of the vertices.

Given paths $\pi_1=v_1v_2 \ldots v_{k}$ and $\pi_2=u_1 u_2 \ldots u_{k}$  in~$G$, we say that $\pi_1$ and $\pi_2$ are \emph{non-colliding} if the following conditions are satisfied 
for all $1\le j< k$:
\begin{enumerate} [label={(\roman*)}, leftmargin=0.5in]
    \item $v_j \neq u_j$ (i.e., no vertex collisions),
    \label{condition:collision} 
    \item $(v_j, v_{j+1}) \neq (u_{j+1}, u_j)$ (i.e., no edge collisions).
    \label{condition:edgeSwap}
\end{enumerate}
If $\pi_1$ and $\pi_2$ are of different lengths, we assume
the agent with the shorter path remains in the target state for collision-checking purposes.\footnote{Changing this to have the agents ``disappear'' (which is a commonly-used assumption) at the target location does not impact our results in any way.}

Given $n$ agents on a graph $G$ and two lists $s_1, \ldots, s_n$ 
and $g_1, \ldots, g_n$ of source and goal vertices, respectively, a \emph{plan} $P=\{\pi_1, \ldots, \pi_n\}$ is a set of 
paths such that $\pi_i$ drives agent~$i$ from $s_i$ to $g_i$ for every $i\in \{1, \ldots, n\}$. A plan is called \emph{non-colliding} if $\pi_i$ and $\pi_j$ are non-colliding for all $i\neq j \in \{1,\ldots,n\}$.
The \emph{length} of the plan, denoted by~$\length(P)$, is the maximal length of a path in~$P$. The \emph{sum-of-costs} (SOC) of~$P$, denoted by $\|P\|$, is the sum of lengths of all the paths in~$P$.
The classical \emph{Multi-Agent Path Finding (MAPF)} problem is to find a non-colliding plan\footnote{Typically, the plan is required to be optimal with respect to some cost function, e.g., length or sum-of-costs.} $P$ in $G$ with the given source and target vertices. 

We now turn to formalize delays and delay-introduction.
Consider a path $\pi=v_1\ldots v_m$ and some $d\in \bbN$. We say that a path $\pi'$ is a \emph{$d$-delay} of $\pi$ if $\pi'=v_1v_1^{k_1}v_2v_2^{k_2}\ldots v_mv_m^{k_m}$, where $v_i^{k_i}$ means repeating $v_i$ for an additional $k_i\ge 0$ times.
That is, $\pi'$ repeats some of the vertices of $\pi$ so that the total amount of repetitions is $d$. 
Note that if $k_i=0$ for all $i\in\{1, \ldots, m\}$ then $\pi'=\pi$, i.e., $\pi'$ is a $0$-delay path of $\pi$.
Also, $\pi$ might already have vertex repetitions (e.g., it could be that $v_1=v_2$). Thus, we only allow adding repetitions, not removing them.

\begin{problem}[Avoiding Collisions by Introducing Delays (\textbf{ACID})]
\label{prob:delay_introduction}
    Given a graph $G=\tup{V,E}$, a plan $P=\{\pi_1,\ldots,\pi_n\}$ and a budget $D\in \bbN$, decide whether there exist paths $\pi'_1,\ldots,\pi'_n$ where $\pi'_i$ is a $d_i$-delay of $\pi_i$ for each $i$, with $\sum_{i=1}^n d_i\le D$ and $P'=\{\pi'_1,\ldots, \pi'_n\}$ is non-colliding.
\end{problem}

Note that ACID is stated as a decision problem, but for algorithmic purposes we consider its optimization variant, in which we want to find a non-colliding plan that minimizes~$D$, i.e., the added length of the plan. 


\section{Computational Complexity of ACID}
\label{sec:delays}
We start our investigation of ACID by establishing its computational complexity. Specifically, we show that solvable instances can be solved using a quadratic  delay.
\begin{lemma}
\label{lem:quadratic_upper_bound}
Consider an ACID instance with plan $P=\{\pi_1,\ldots,\pi_n\}$ and budget $D$. If the instance is solvable, then it is also solvable with budget $D'=(n-1) \cdot \|P\|$.
\end{lemma}
\begin{proof}
    Intuitively, the maximal number of delays we may need to introduce is such that we ``spread'' $P$ so that only a single agent moves at each timestep, and the remaining $n-1$ agents are delayed. 
    
    Formally, consider a plan $P'$ that is a solution to Problem~\ref{prob:delay_introduction}.
    If there is a time where all agents are delayed simultaneously in $P'$, this delay can be safely removed. Thus, at each step, at least one path advances, so to obtain $P'$ from $P$, we introduce, for each agent and each step of $P$, at most $n-1$ delays.
    Since $P$ comprises a total of $\|P\|$ agent steps, the total delays introduced are at most $(n-1) \cdot \|P\|$. 
\end{proof}
\begin{remark}[Encoding of the budget $D$]
\label{rmk:encoding_budget}
    ACID can be considered with $D$ encoded either in binary or in unary. We note that this does not affect the computational complexity, since by \cref{lem:quadratic_upper_bound}, we can assume w.l.o.g. that $D$ is polynomially bounded in the size of $P$ (namely in $n$ and $\|P\|$).
\end{remark}
We are now ready to establish the complexity of ACID. The complete proof can be found in 
Section~\ref{sec:appendix}.
For background on hardness of approximation see~\cite{APX-background}.

\begin{theorem}
\label{thm:ACID_NP_complete}
ACID is \NP-complete, and even \APX-hard.
\end{theorem}
\begin{proof}[Proof sketch]
Membership in $\NP$ follows immediately from \cref{lem:quadratic_upper_bound}---simply guess a set of (polynomially bounded) delays and check that the resulting plan is non-colliding.

We turn to show $\NP$-hardness by showing a reduction from the Minimum Sum Coloring (MSC) problem, defined as follows. In an MSC instance we are given an undirected graph $G=\tup{V,E}$ and a threshold $C$ (encoded in unary), and the goal is to decide whether there exists a coloring $\chi:V\to \bbN$ such that $\sum_{v\in V}\chi(v)\le C$. That is, we need to color the vertices of $G$ with natural numbers (including $0$), such that every two vertices that share an edge are assigned different colors, and the sum of colors is at most $C$. MSC was shown to be \NP-complete in~\citet{SumColoring} and 
is \APX-hard~\cite{SumColoring-APX}.

We start with an intuitive overview of the reduction and depict the reduction in~\cref{fig:graph_reduction,fig:output_reduction}.
Given a graph $G=\tup{V,E}$ and $C\in \bbN$, denote $V=\{1,\ldots, n\}$ and $E=\{e_1,\ldots, e_m\}$. Our ACID instance consists of $n$ agents travelling along a concatenation of $C+1$ identical \emph{blocks}, constructed as follows. 
For agent $i\in \{1,\ldots n\}$ we build a path of length $m$ that for the most part is disjoint from all other paths. However, for each edge $e_r\in E$, if $e_r=\{i,j\}$ for some $j\in V$, then node $r$ in the path is shared by the paths of agents $i$ and $j$. For example, edge $e_2$ in~\cref{fig:graph_reduction} appears on both the green and red paths in the blocks in~\cref{fig:output_reduction}.
Each agent starts its traversal from a distinct initial node, and the blocks are concatenated in the natural way. Note that since $C$ is in unary, the reduction is polynomial.

We claim that $G$ can be colored with sum at most $C$ if and only if the resulting ACID instance has a solution with budget $C$.
Intuitively, without introducing delays, for every edge $e_r=\{i,j\}\in E$ agents $i$ and $j$ collide in each block on the node corresponding to $e_r$. However, if $i$ and $j$ are delayed by different amounts before reaching node $e_r$, then they do not collide. 

Thus, one direction of the proof is easy: if $G$ has a coloring $\chi$ of sum at most $C$, then by delaying agent $i\in V$ for $\chi(i)$ steps (hence keeping within the budget $C$), the agents do not collide in any block. 

The converse direction is more involved. 
Assume the ACID instance has a solution with budget at most $C$. Since there are $C+1$ blocks, it follows that there is at least one block where the agents are not delayed. In the technical appendix we show that we can therefore assume all following blocks contain no delays as well, and moreover -- that we can aggregate all the delays to the initial node of each agent. These delays induce a coloring of $G$ of sum at most $C$, whereby each agent is colored with its number of delays. Since the agents do not collide, this coloring is legal.
The hardness of approximation follows from the fact that our reduction preserves the cost of a solution.
\end{proof}
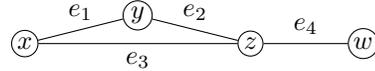
\begin{figure}[t]
    \centering
        \begin{tikzpicture}[scale=0.75, 
nodestyle/.style={circle,draw,inner sep=0pt,minimum size=3pt,fill=black}, 
xlinestyle/.style={blue,-stealth}, 
ylinestyle/.style={ForestGreen,-Latex}, 
zlinestyle/.style={Red,-Triangle}, 
wlinestyle/.style={RoyalPurple,-Kite},
nodelabel/.style={draw=}]
	
        \node[draw,circle, inner sep=1pt, minimum size=3pt,black] (x) at (0,0) {$x$};
        \node[draw,circle, inner sep=1pt, minimum size=3pt,black] (y) at (2,0.5) {$y$};
        \node[draw,circle, inner sep=1pt, minimum size=3pt,black] (z) at (4,0) {$z$};
        \node[draw,circle, inner sep=1pt, minimum size=3pt,black] (w) at (6,0) {$w$};

        \draw (x) edge node[above] {$e_1$} (y);
        \draw (y) edge node[above] {$e_2$} (z);
        \draw (z) edge node[below] {$e_3$}(x);
        \draw (z) edge node[above] {$e_4$} (w);

\end{tikzpicture}
    \caption{An input graph for the reduction with $C=3$. Observe that the graph can be colored with sum $3$, by $\chi(x)=\chi(w)=0$, $\chi(y)=1$ and $\chi(z)=2$.
    Note that for clarity, we use $x,y,z$ and $w$ and not $1, \ldots, 4$ (as is done in the reduction) to name the vertices.
    }
    \label{fig:graph_reduction}
\end{figure}
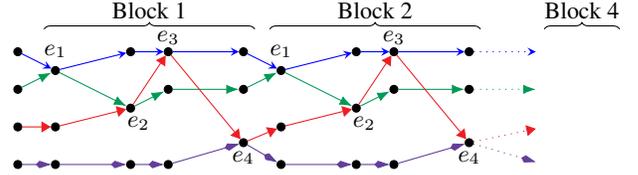
\begin{figure}[t]
    \centering
    \begin{tikzpicture}[scale=0.5, 
nodestyle/.style={circle,draw,inner sep=0pt,minimum size=3pt,fill=black}, 
xlinestyle/.style={blue,-stealth}, 
ylinestyle/.style={ForestGreen,-Latex}, 
zlinestyle/.style={Red,-Triangle}, 
wlinestyle/.style={RoyalPurple,-Kite},
nodelabel/.style={draw=}]
	\footnotesize
	\foreach \d in {0,...,3}{
		\node[nodestyle] (T\d0) at (0,-\d) {};
	}

        \foreach \d in {0,...,1}{
            \pgfmathtruncatemacro{\xcor}{\d*6+1.5};
            \pgfmathtruncatemacro{\ycor}{\d*6+3};
            \pgfmathtruncatemacro{\zcor}{\d*6+4.5};
            \pgfmathtruncatemacro{\wcor}{\d*6+6};
            
            \node[nodestyle,label={[xshift=0pt]$e_1$}] (XY1\d) at (\xcor,-0.5) {};
            \node[nodestyle] (Z1\d) at (\xcor,-2) {};
            \node[nodestyle] (W1\d) at (\xcor,-3) {};

            \node[nodestyle] (X2\d) at (\ycor,0) {};
            \node[nodestyle,label={[yshift=-13pt,xshift=3pt]$e_2$}] (YZ2\d) at (\ycor,-1.5) {};
            \node[nodestyle] (W2\d) at (\ycor,-3) {};

            \node[nodestyle,label={[yshift=-2pt]$e_3$}] (XZ3\d) at (\zcor,0) {};
            \node[nodestyle] (Y3\d) at (\zcor,-1) {};
            \node[nodestyle] (W3\d) at (\zcor,-3) {};

            \node[nodestyle] (X4\d) at (\wcor,0) {};
            \node[nodestyle] (Y4\d) at (\wcor,-1) {};
            \node[nodestyle,label={[yshift=-13pt]$e_4$},below] (ZW4\d) at (\wcor,-2.3) {};

             \draw [xlinestyle] (XY1\d) edge (X2\d) (X2\d) edge (XZ3\d) (XZ3\d) edge (X4\d); 
             \draw [ylinestyle,ForestGreen] (XY1\d) edge (YZ2\d) (YZ2\d) edge (Y3\d) (Y3\d) edge (Y4\d);
             \draw [zlinestyle,Red] (Z1\d) edge  (YZ2\d) (YZ2\d) edge (XZ3\d) (XZ3\d) edge (ZW4\d);
             \draw [wlinestyle,RoyalPurple] (W1\d) edge (W2\d) (W2\d) edge (W3\d) (W3\d) edge (ZW4\d);
        }
        \draw [xlinestyle] (T00) -- (XY10); 
        \draw [xlinestyle] (X40) -- (XY11);
        \node (X_END) at (14,0){}; 
        \draw [xlinestyle,dotted] (X41) -- (X_END);
        \draw [ylinestyle,ForestGreen] (T10)--(XY10);
        \draw [ylinestyle,ForestGreen] (Y40)--(XY11);
        \node (Y_END) at (14,-1){}; 
        \draw [ylinestyle,ForestGreen,dotted] (Y41) -- (Y_END);
        \draw [zlinestyle,Red] (T20)-- (Z10);
        \draw [zlinestyle,Red] (ZW40)-- (Z11);
        \node (Z_END) at (14,-2){}; 
        \draw [zlinestyle,Red,dotted] (ZW41) -- (Z_END);
        \draw [wlinestyle,RoyalPurple] (T30)--(W10);
        \draw [wlinestyle,RoyalPurple] (ZW40)--(W11);
        \node (W_END) at (14,-3){}; 
         \draw [wlinestyle,RoyalPurple,dotted] (ZW41) -- (W_END);
        
        \draw [decorate, decoration = {brace}] (0.7,0.6) --  (6.3,0.6);
        \draw [decorate, decoration = {brace}] (6.7,0.6) --  (12.3,0.6);
        \draw [decorate, decoration = {brace}] (14,0.6) --  (16,0.6);

        \node at (3.5,1.1) {Block 1};
        \node at (9.5,1.1) {Block 2};
        \node at (15,1.1) {Block 4};


 
        
       
\end{tikzpicture}
    \caption{Reduction output. Each agent is represented by a path (e.g., $x$ is the blue path, also distinguished by arrow types). The complete output has $C+1=4$ blocks.
    }
    \label{fig:output_reduction}
\end{figure}

\begin{remark}[ACID variants]
    \label{rmk:variants_of_ACID}
    The hardness in \cref{thm:ACID_NP_complete} holds already for the most ``relaxed'' version of ACID. However, the upper bound still holds with various restrictions on the delays, such as only allowing a certain number (possibly zero) of delays per agent, or per node. Thus, ACID remains \NP-complete even if the agents are not allowed to delay in some nodes, or if the budget is specified for each agent, or most generally -- if each agent-vertex pair has a budget.
\end{remark}

\section{MAPF Formulation of ACID}
We turn our attention to developing an algorithmic approach for solving ACID.
To this end, we reduce ACID to a version of MAPF, and utilize existing solutions for the latter. Crucially, we show that the specific MAPF instances resulting from our reduction have certain favorable properties which render them amenable to scalable optimal solutions.

Before detailing our approach, we present a small modification to the MAPF problem, whereby we allow a different set of edges for each agent.
An instance of \emph{MAPF with agent-specific edges} (dubbed \emph{Agent-Edge MAPF}) is a set of vertices $V$ and sets $E_1,\ldots,E_n\subseteq V\times V$ of edges, as well as start and goal vertices for each of the $n$ agents.
The remaining definitions are identical to MAPF, with the exception that a path for agent $i$ must use only edges from $E_i$.

From an algorithmic perspective, solving Agent-Edge MAPF is similar to solving MAPF, in the following sense.
\begin{observation}
\label{obs:agent-edge MAPF}
An algorithm $\mathcal{A}$ for MAPF whose queries to the graph are only stated in the form ``what are the edges from vertex $v$ for agent $i$?'' can solve Agent-Edge MAPF and preserve the same optimality/bounded sub-optimality/anytime properties of the original algorithm~$\mathcal{A}$.
\end{observation}

Note that
(i)~if $\mathcal{A}$ makes only such queries, it cannot distinguish between a (regular) graph and the agent-specific edge setting
and that
(ii)~most common MAPF solvers such as all $A^*$-based solvers like CBS~\cite{sharon2015conflict} and PBS~\cite{ma2019searching} satisfy the condition of~\cref{obs:agent-edge MAPF}. 




\subsection{Constrained Graph}
\label{subsec:CG}
Our reduction of ACID to Agent-Edge MAPF is as follows. Consider an ACID instance with graph $G=\tup{V,E}$ and a plan $P=\{\pi_1,\ldots,\pi_n\}$ (ignore the budget for now). We construct an Agent-Edge MAPF instance with the vertices $V\times \{1,\ldots,\length(P)\}$ (i.e., a copy of $V$ for each step in~$P$, up to the longest path), and the edges are defined by the paths in~$P$, as well as self-loops. 
That is, let $\pi_i=v^i_1,\ldots,v^i_k$. Then we define 
$E_i=\{((v^i_j,j),(v^i_{j+1},j+1))\mid 1\le j< k\}\cup \{((v^i_j,j),(v^i_j,j))\mid 1\le j<k\}$.
We set the start and goal vertices for agent $i$ as $(v^i_1,1)$ and $(v^i_k,k)$, respectively.

\begin{figure}[t]
    \centering
    \begin{subfigure}{0.49\columnwidth}
        \centering
        \includegraphics[width=\columnwidth]{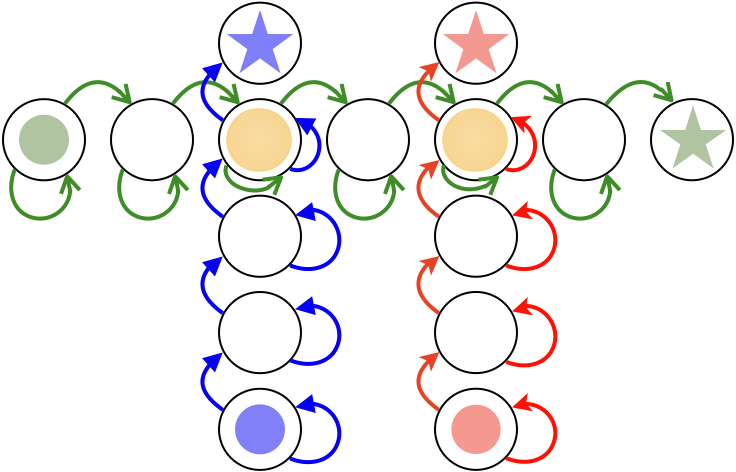}
        \caption{Constrained Graph.}
        \label{fig:constrained_graph}
    \end{subfigure}
    \hfill
    \begin{subfigure}{0.5\columnwidth}
        \centering
        \includegraphics[width=\columnwidth]{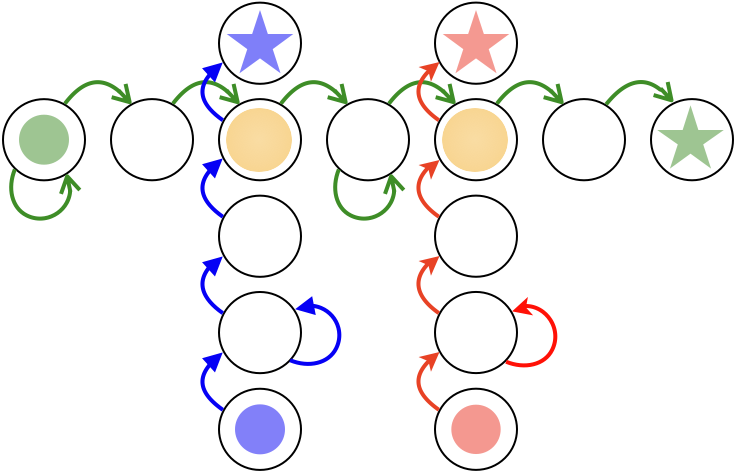}
        \caption{Improved Constrained Graph.}
        \label{fig:improved_constrained_graph}
    \end{subfigure}
    \caption{(a) A Constrained Graph and (b) an Improved Constrained Graph for an MAPF plan with three agents.
    }
    \label{fig:cg_and_icg}
\end{figure}

We refer to the multiple-edgeset graph obtained above as the \emph{Constrained Graph (CG)} (see \cref{fig:constrained_graph}). Notice that the Agent-Edge graph does not allow agents to deviate from their original paths.
That is, the green agent located at either highlighted vertex must either delay or transition to the next immediate right vertex. Similarly, the red and blue agents located at the same vertices must either delay or move upward.

Note that the out-degree (number of outgoing edges) of each vertex in a CG is at most two. This makes a CG an ``easy'' candidate for planning since the branching factor used by search algorithms is often (though not necessarily) small, implying low running times.

We later show that an ACID instance has a solution with delay $D$ iff the Agent-Edge MAPF instance on the CG has a solution of length $\|P\|+D$. But first, we present an optimization over the CG which further eliminates redundancy.



\subsection{Improved Constrained Graph}
\label{subsec:ICG}
Observe that the CG has self loops on all 
vertices. This, however, may be redundant. For example, the collision from Fig.~\ref{fig:example_delay_late} can be resolved by having the red agent delay at \emph{any} of the vertices along its path prior to the conflict. Thus, it suffices to have a self loop on only one of the vertices along its path prior to the conflict  (see \cref{fig:improved_constrained_graph}).


Given a CG, we construct an \emph{Improved} CG (ICG) as follows. For each path $\pi_i=v^i_1,\ldots,v^i_k$ in the plan, let $I_i\subseteq \{1,\ldots, k\}$ be the set of indices $j$ for which $v^i_j$ occurs also in some other path $\pi_l$. We refer to $I_i$ as the \emph{intersecting indices} of agent $i$.
The ICG is an Agent-Edge MAPF instance obtained from a CG so that between every two intersecting indices we keep exactly one vertex with a self-loop.
Formally, for every agent $i$ and $s<t\in I_i\cup \{0\}$
we retain one self loop in the vertices $v^i_{s+1}\ldots v^i_t$.

Note that computing ICG from CG is easy---we simply find the intersecting vertices in quadratic time, and scan the intervals between them.

As we prove in Section~\ref{sec:appendix},
using ICG instead of CG is sound and complete, in the following sense.
\begin{theorem}
\label{thm:ACID_equiv_CG_equiv_ICG}
Given an instance of ACID with plan $P$, let the CG and ICG be as above. Then, the following are equivalent.
\begin{enumerate}
    \item The ACID instance has a solution with budget $D$.
    \item The Agent-Edge MAPF of CG has a solution $P'$ with $\|P'\|\le \|P\|+D$.
    \item The Agent-Edge MAPF of ICG has a solution $P'$ with $\|P'\|\le \|P\|+D$.
\end{enumerate}
\end{theorem}
Following~\cref{thm:ACID_equiv_CG_equiv_ICG}, we can solve an ACID instance by applying \emph{any} MAPF algorithm that satisfies \cref{obs:agent-edge MAPF} to the corresponding CG or ICG.
\section{Experimental Evaluation}
\label{sec:experiments}
\begin{figure*}[t]
     \centering
     \begin{subfigure}{0.1\textwidth}
         \centering
         \includegraphics[width=\textwidth]{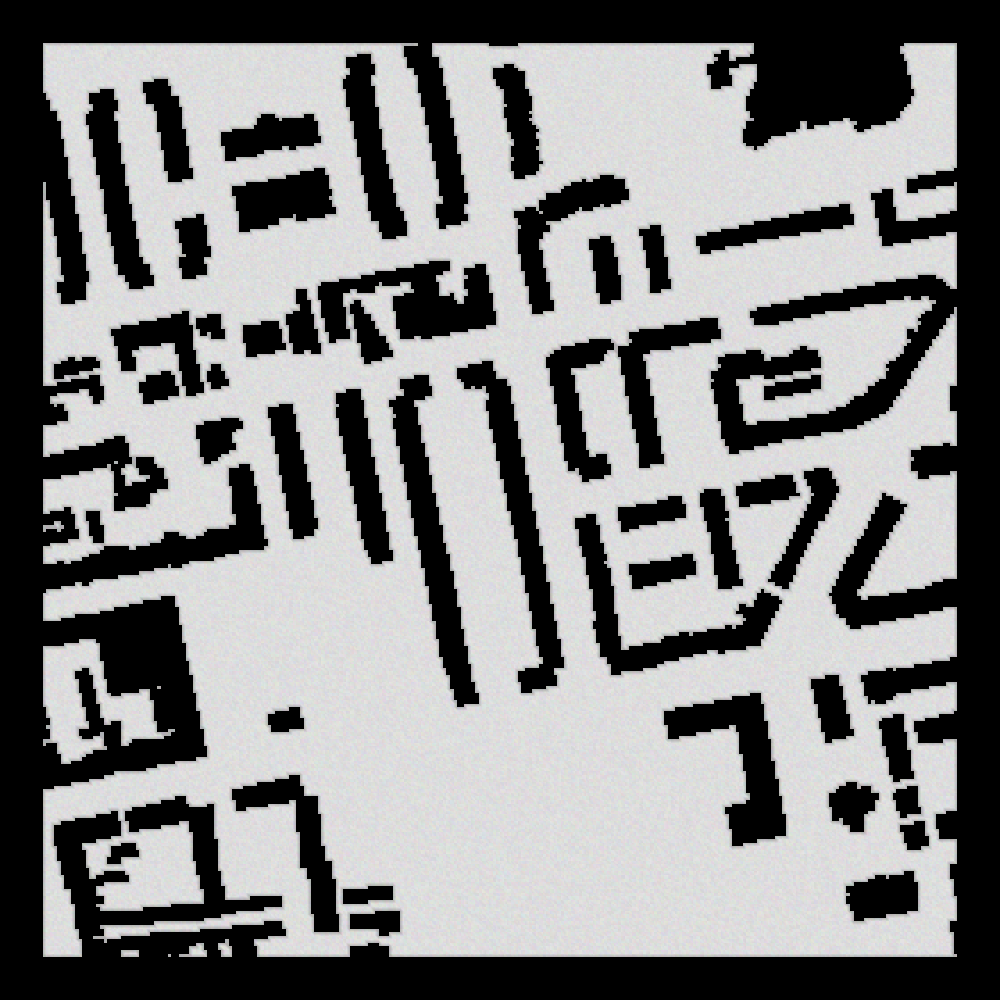}
         \caption{Berlin}
         \label{fig:Berlin_1_256}
     \end{subfigure}
     \hfill
     \begin{subfigure}{0.1\textwidth}
         \centering
         \includegraphics[width=\textwidth]{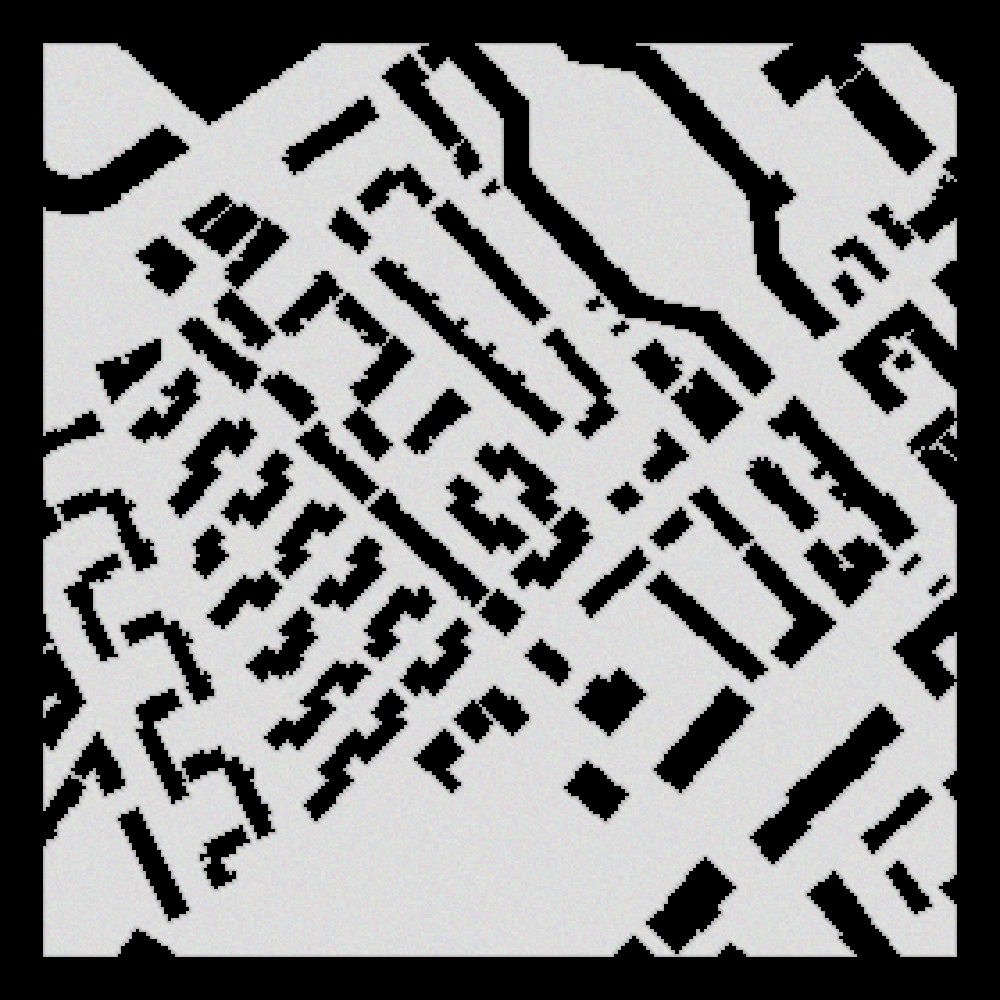}
         \caption{Boston}
         \label{fig:Boston_0_256}
     \end{subfigure}
     \hfill
     \begin{subfigure}{0.1\textwidth}
         \centering
         \includegraphics[width=\textwidth]{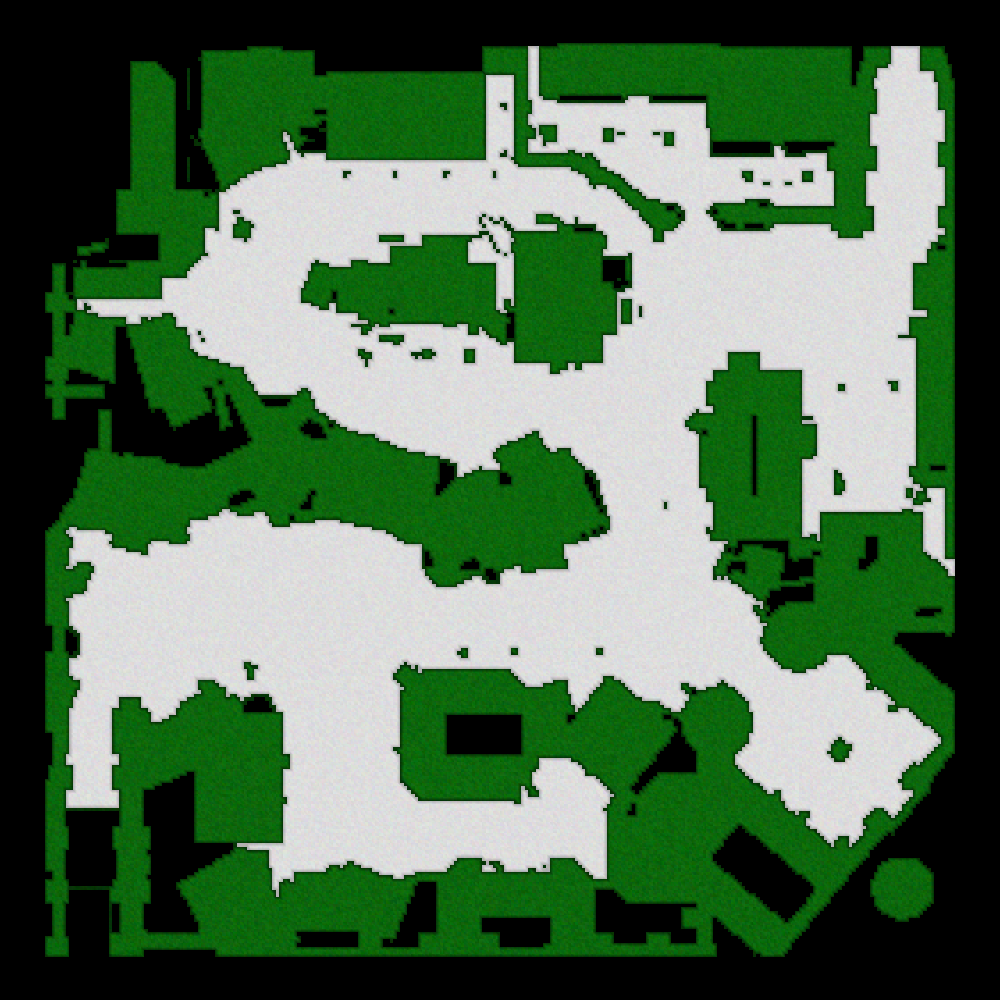}
         \caption{den520d}
         \label{fig:den520d}
     \end{subfigure}
     \hfill
     \begin{subfigure}{0.1\textwidth}
         \centering
         \includegraphics[width=\textwidth]{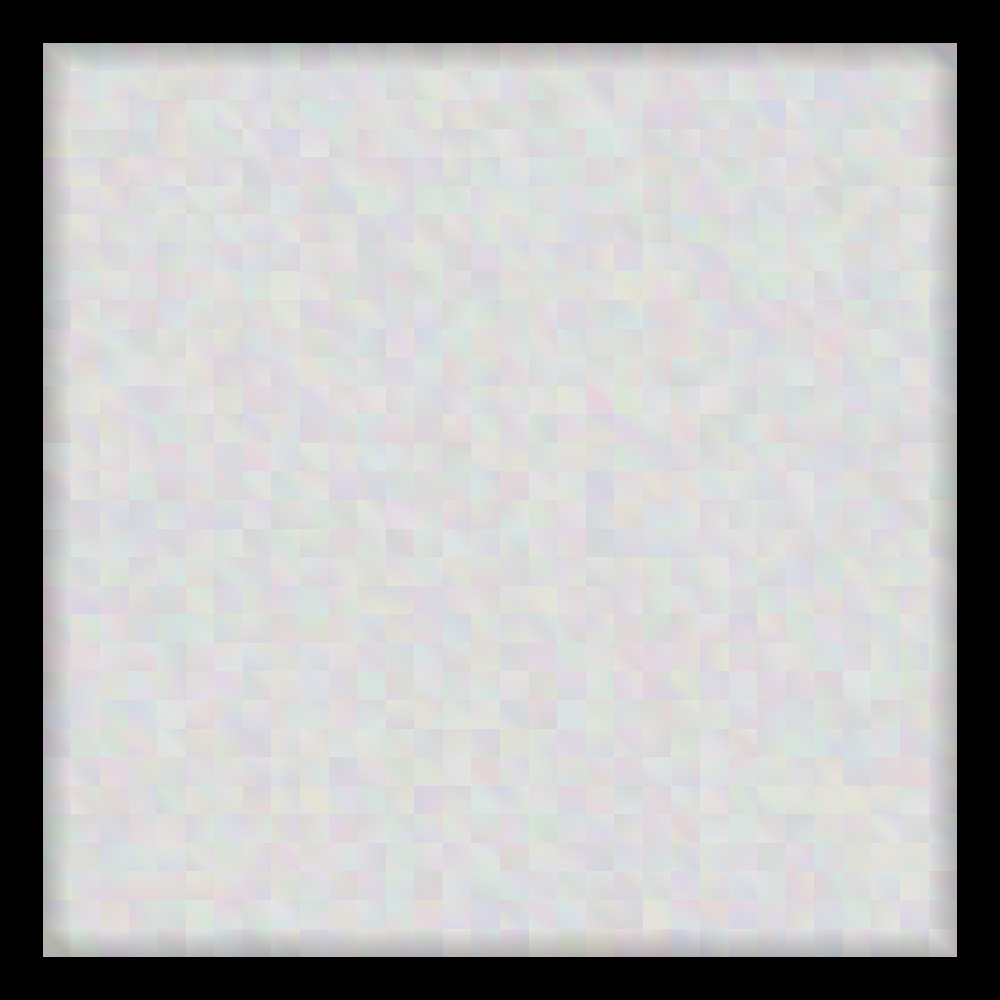}
         \caption{empty $32$}
         \label{fig:empty-32-32}
     \end{subfigure}
     \hfill
     \begin{subfigure}{0.1\textwidth}
         \centering
         \includegraphics[width=\textwidth]{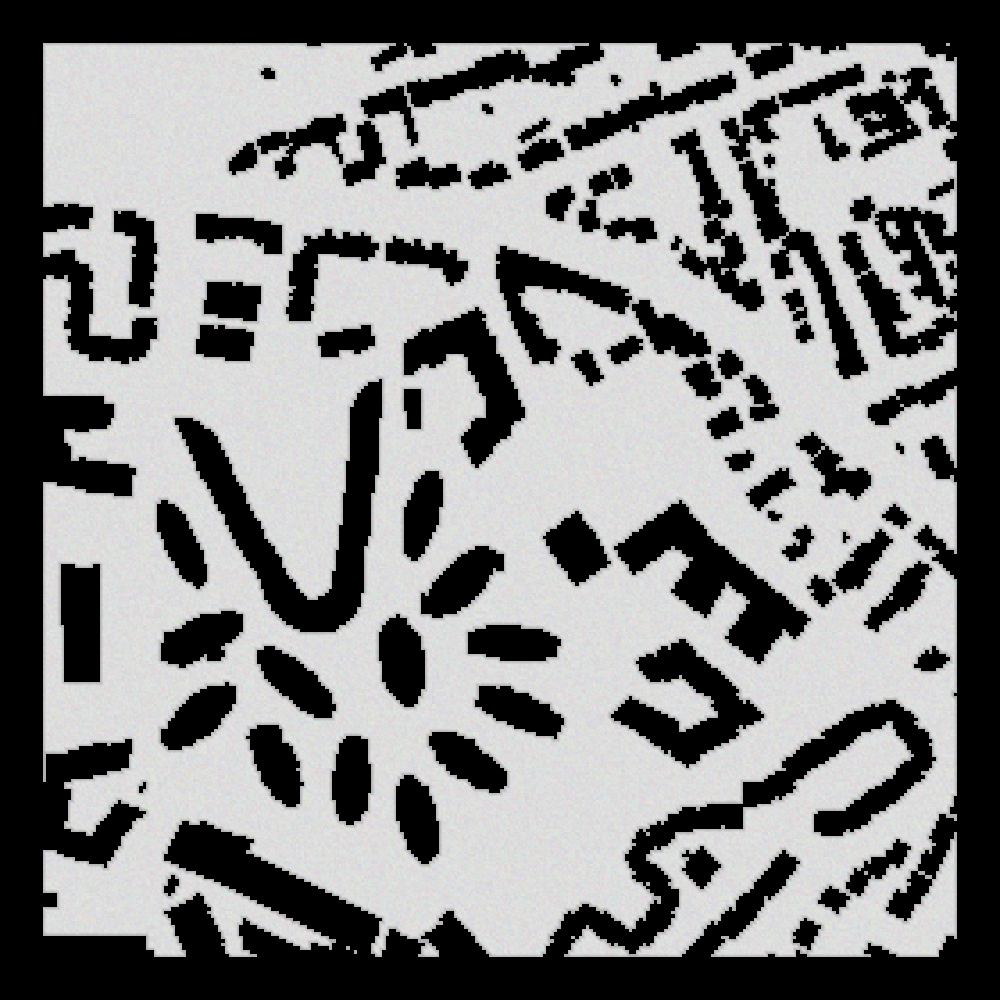}
         \caption{Paris}
         \label{fig:Paris_1_256}
     \end{subfigure}
     \hfill
     \begin{subfigure}{0.1\textwidth}
         \centering
         \includegraphics[width=\textwidth]{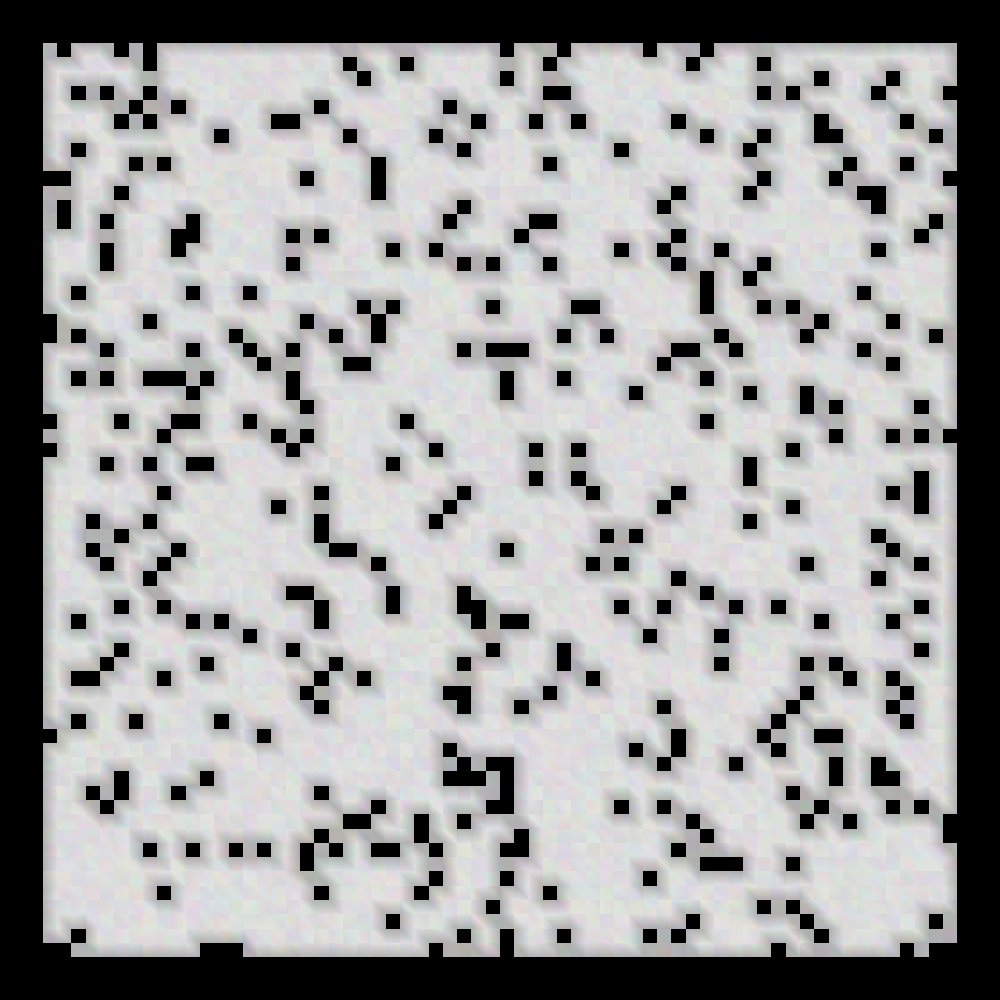}
         \caption{random $64$}
         \label{fig:random-64-64-10}
     \end{subfigure}
     \hfill
     \begin{subfigure}{0.1\textwidth}
         \centering
         \includegraphics[width=\textwidth]{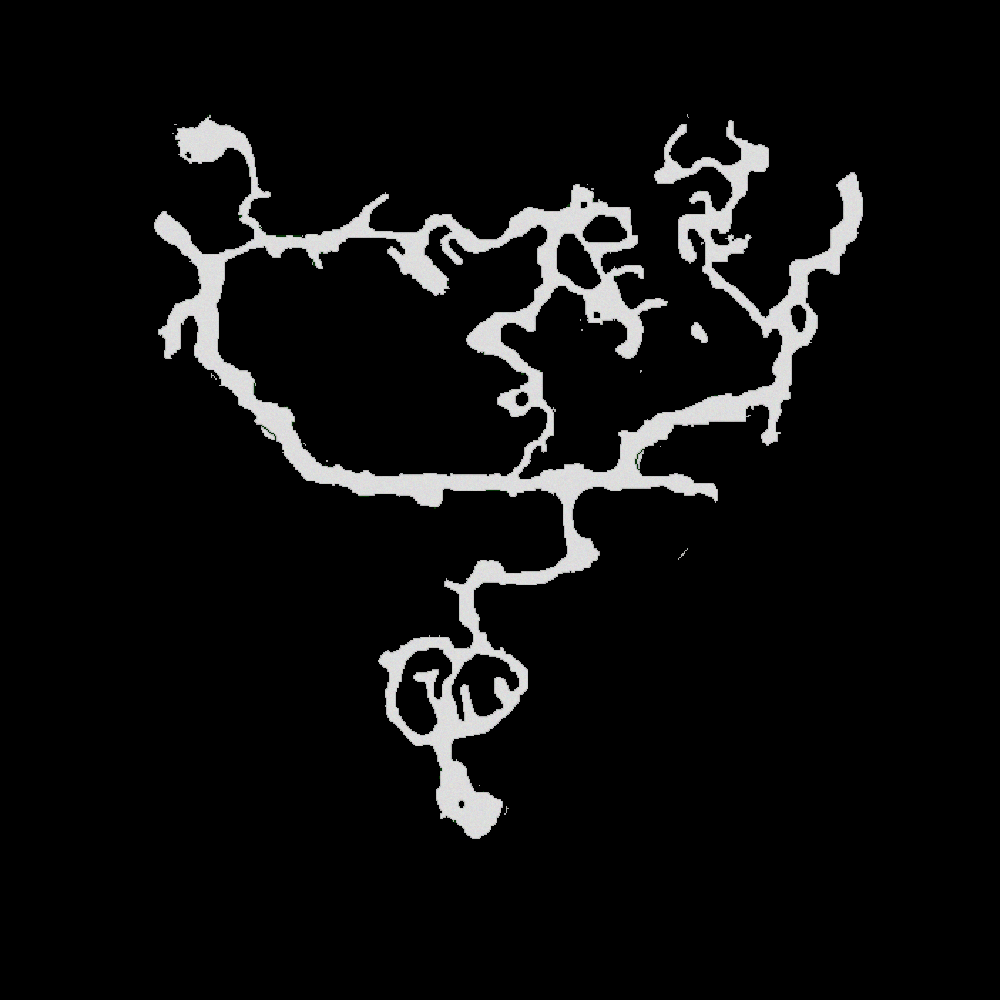}
         \caption{w-coast}
         \label{fig:w_woundedcoast}
     \end{subfigure}
     \hfill
     \begin{subfigure}{0.115\textwidth}
         \centering
         \includegraphics[width=.88\textwidth]{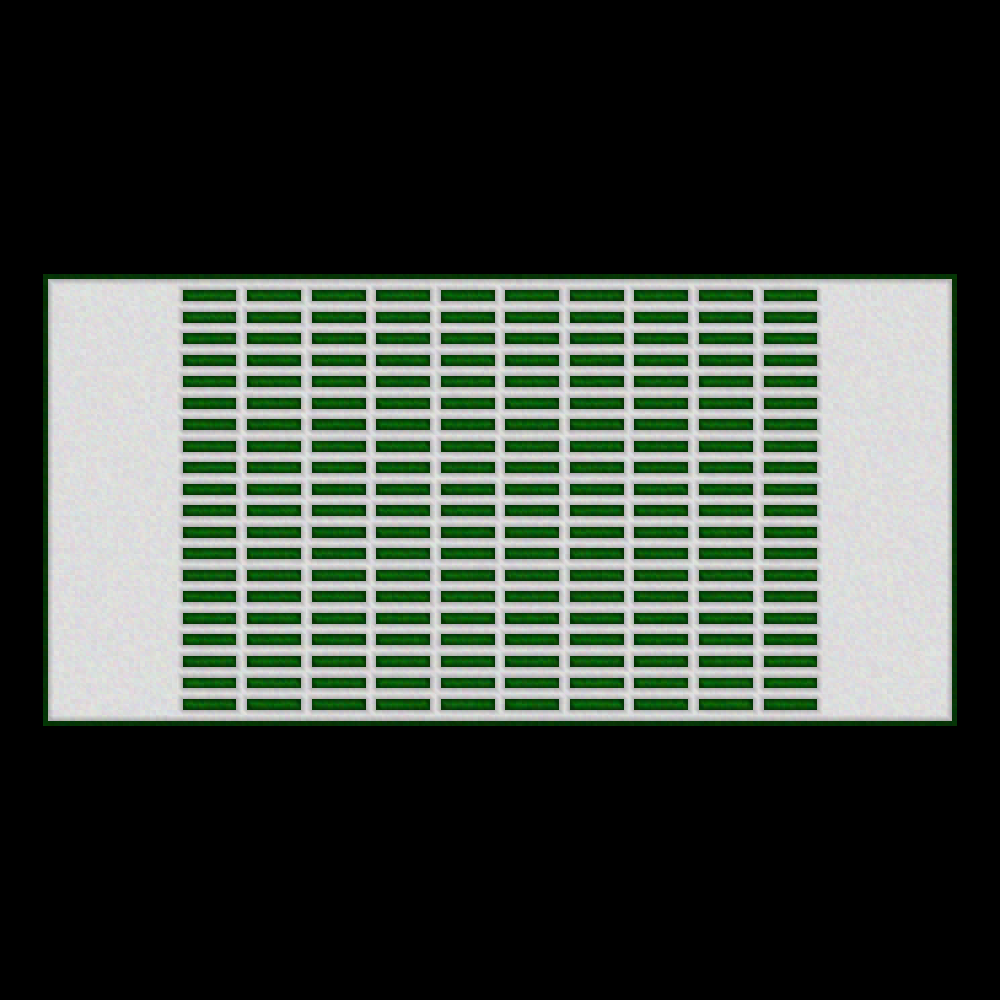}
         \caption{warehouse-1}
         \label{fig:warehouse-10-20-10-2-2}
     \end{subfigure}
     \hfill
     \begin{subfigure}{0.115\textwidth}
         \centering
         \includegraphics[width=.88\textwidth]{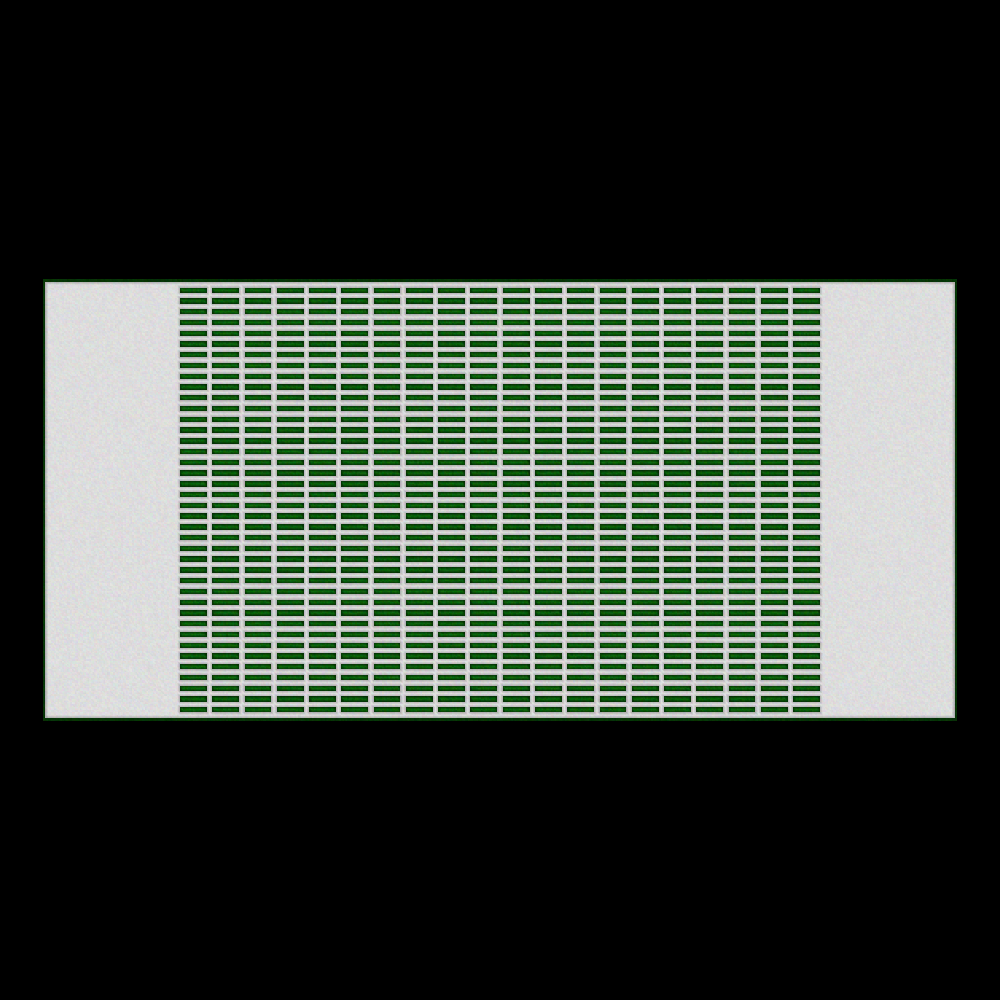}
         \caption{\small{warehouse-2}}
         \label{fig:warehouse-20-40-10-2-2}
     \end{subfigure}
        \caption{MAPF maps used in the experimental evaluations.}
        \label{fig:mapf-maps}
\end{figure*}
\begin{figure*}[t!]
    \centering
     \begin{subfigure}{0.3\textwidth}
         \centering
         \includegraphics[width=0.9\textwidth]{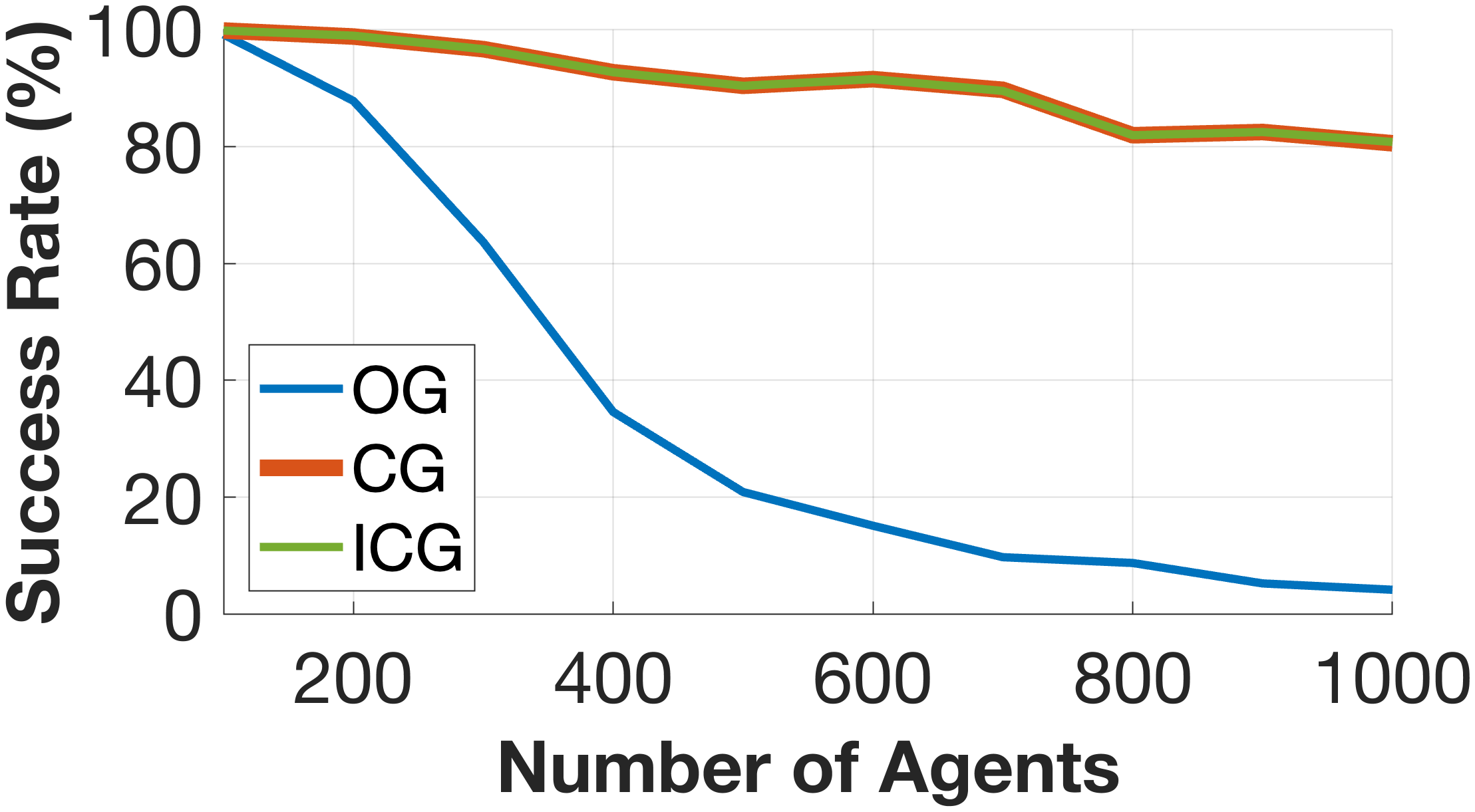}
         \caption{CBS success rates}
         \label{fig:global-cbs-succ}
     \end{subfigure}
     \hfill
     \begin{subfigure}{0.3\textwidth}
         \centering
         \includegraphics[width=0.9\textwidth]{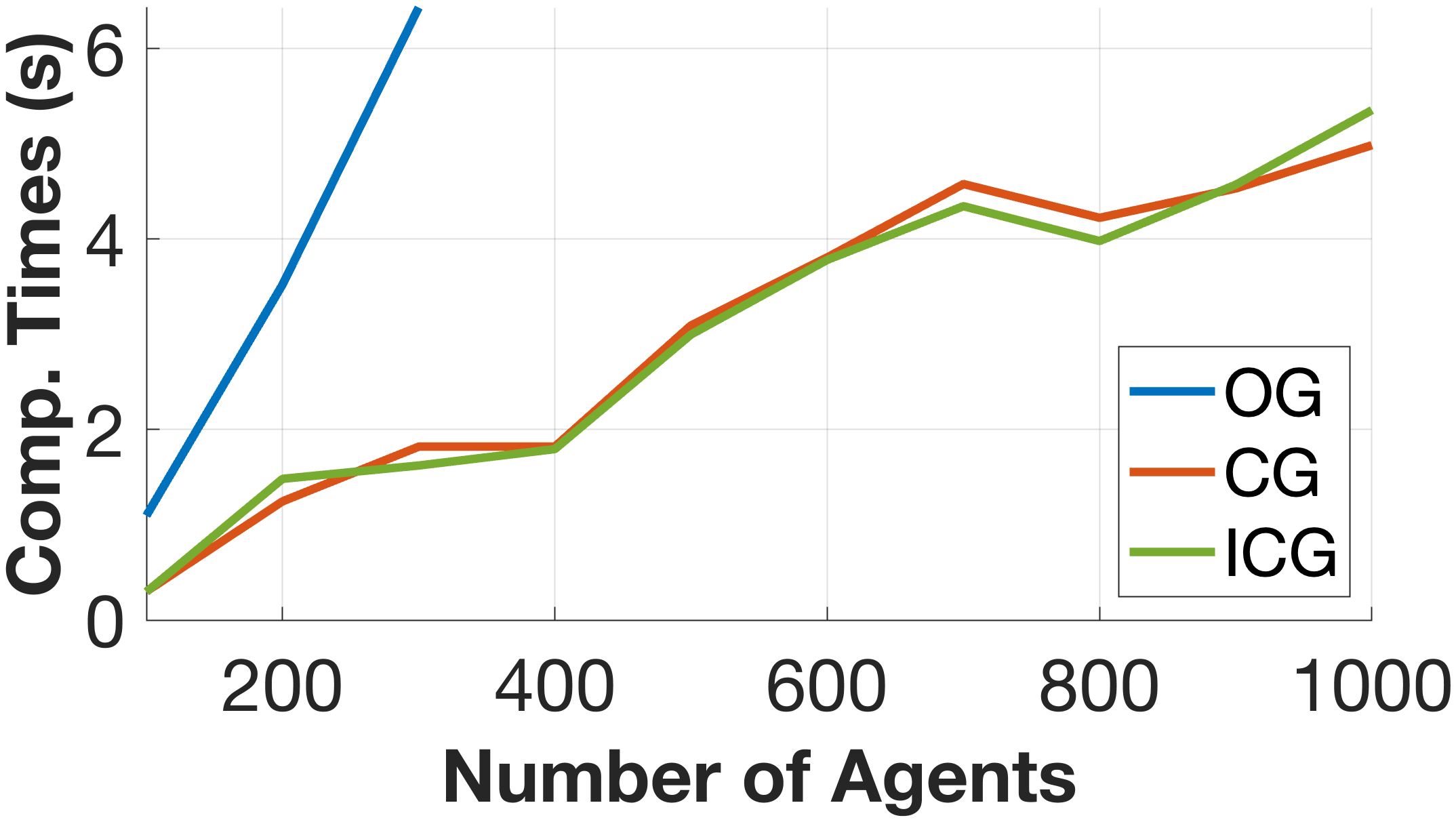}
         \caption{CBS computation times}
         \label{fig:global-cbs-comps}
     \end{subfigure}
     \hfill
     \begin{subfigure}{0.3\textwidth}
         \centering
         \includegraphics[width=0.9\textwidth]{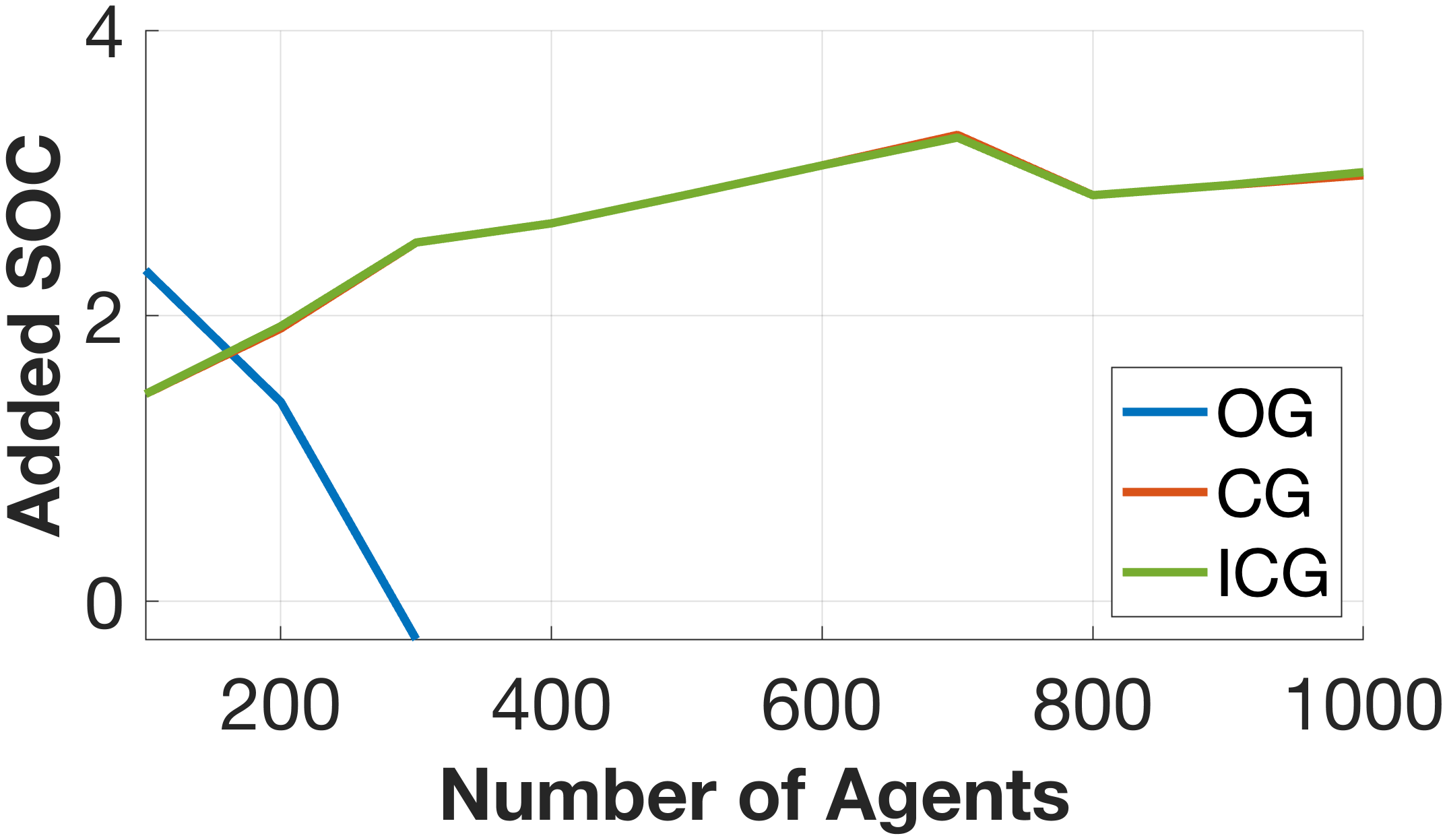}
         \caption{CBS added SOC}
         \label{fig:global-cbs-apl}
     \end{subfigure}
     \newline
     \begin{subfigure}{0.3\textwidth}
         \centering
         \includegraphics[width=0.9\textwidth]{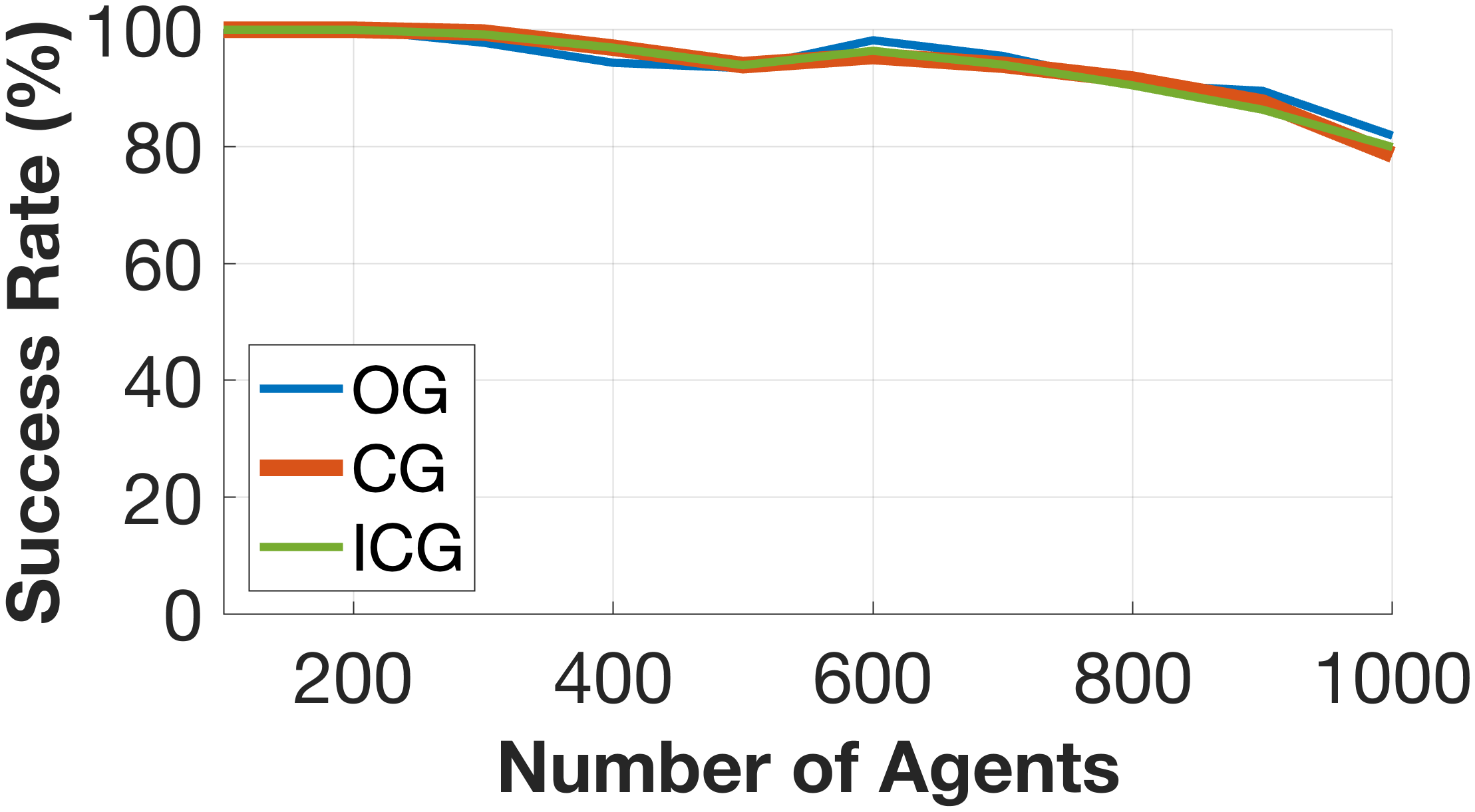}
         \caption{Anytime-EECBS success rates}
         \label{fig:global-eecbs-succ}
     \end{subfigure}
     \hfill
     \begin{subfigure}{0.3\textwidth}
         \centering
         \includegraphics[width=0.9\textwidth]{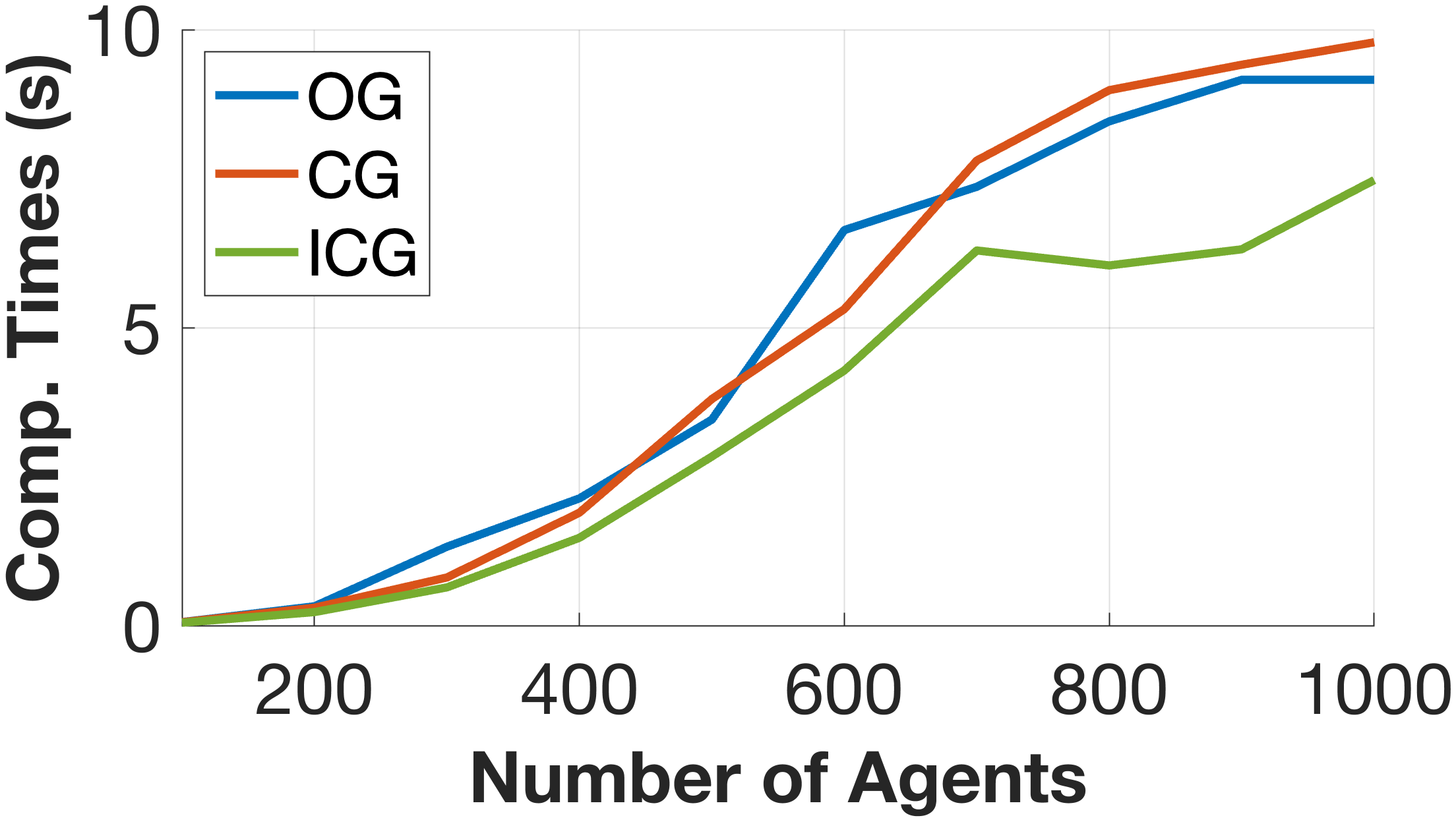}
         \caption{Anytime-EECBS computation times}
         \label{fig:global-eecbs-comps}
     \end{subfigure}
     \hfill
     \begin{subfigure}{0.3\textwidth}
         \centering
         \includegraphics[width=0.9\textwidth]{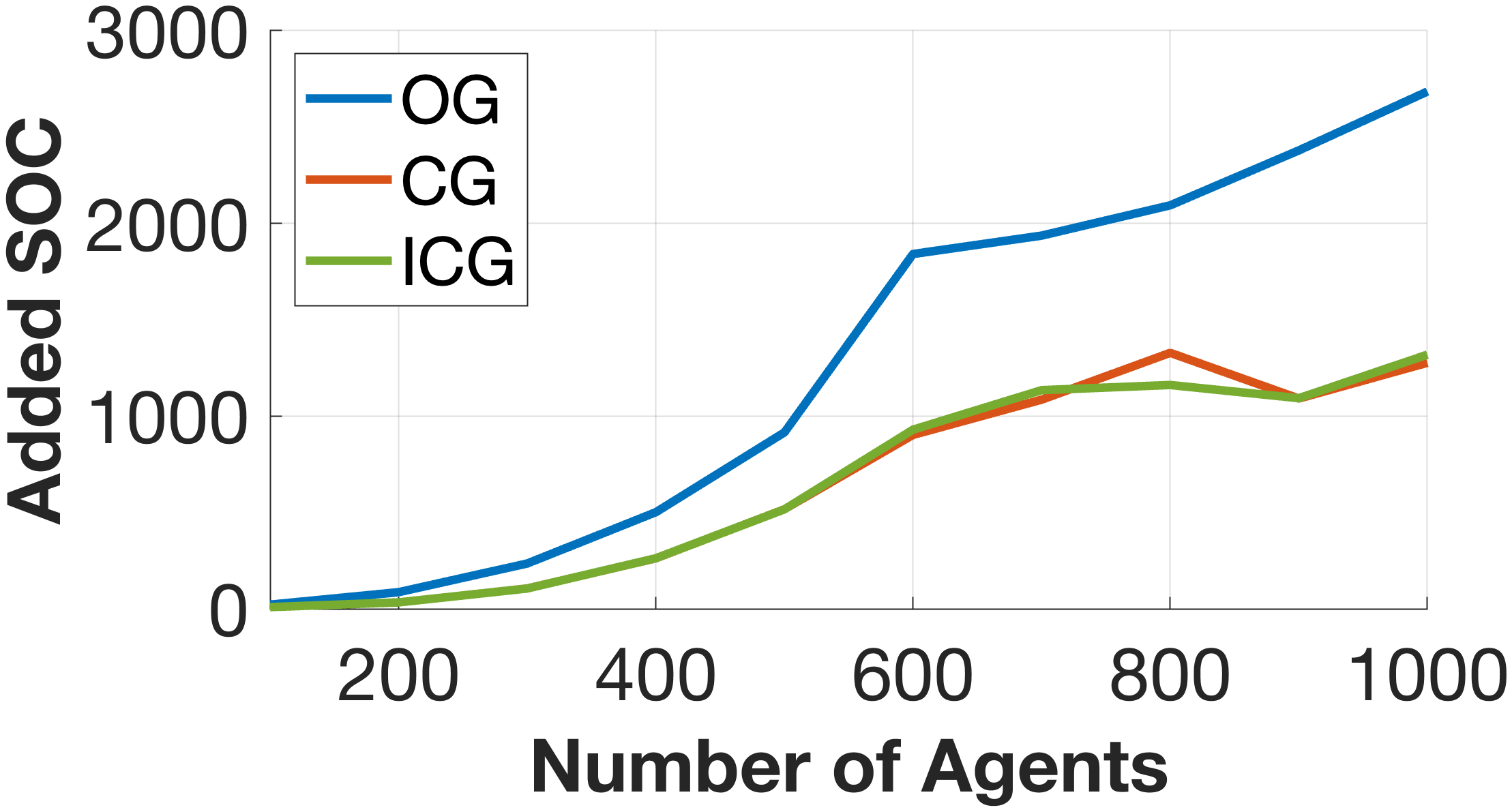}
         \caption{Anytime-EECBS added SOC}
         \label{fig:global-eecbs-apl}
     \end{subfigure}
     \newline
     \begin{subfigure}{0.3\textwidth}
         \centering
         \includegraphics[width=0.9\textwidth]{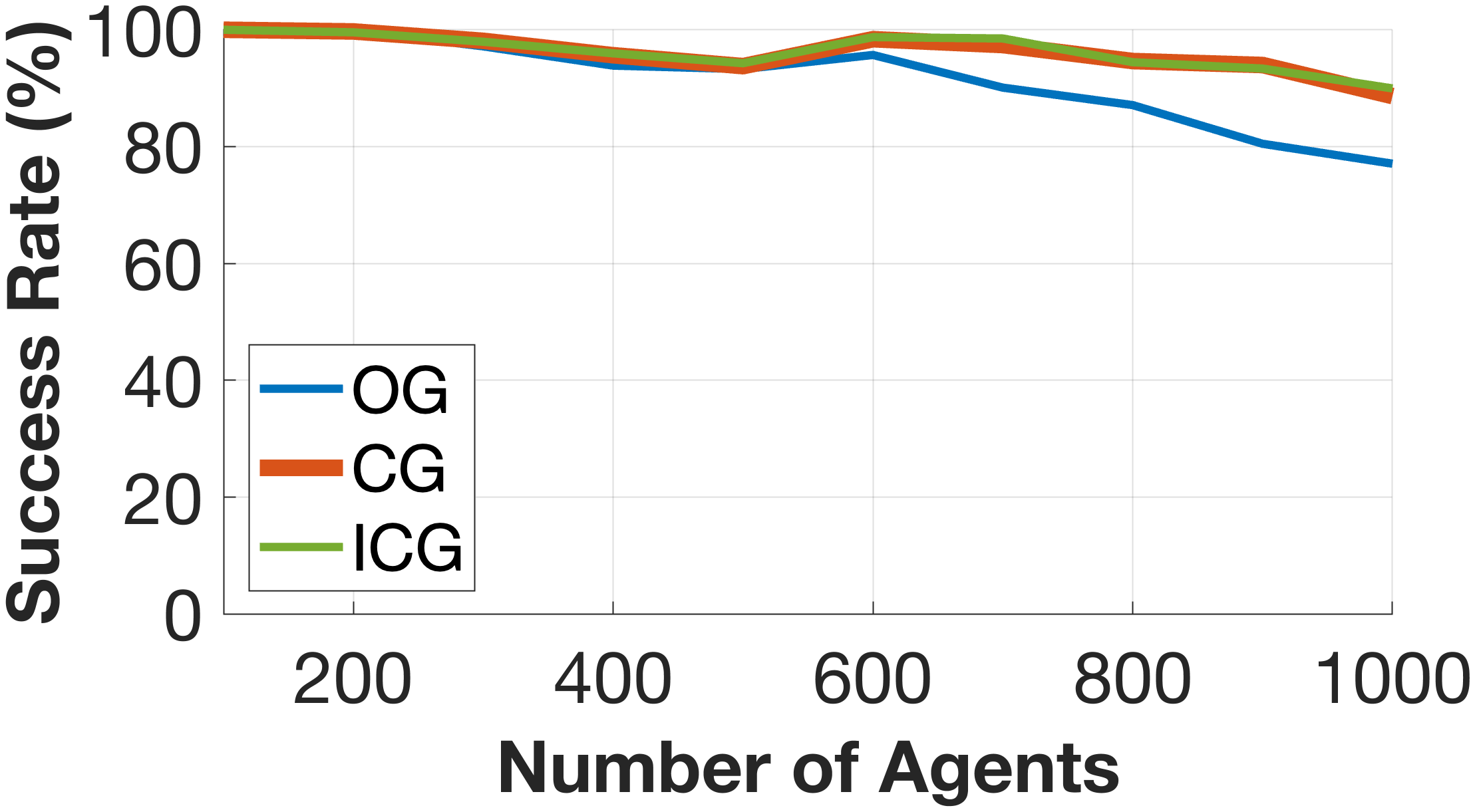}
         \caption{MAPF-LNS2 success rates}
         \label{fig:global-lns-succ}
     \end{subfigure}
     \hfill
     \begin{subfigure}{0.3\textwidth}
         \centering
         \includegraphics[width=0.9\textwidth]{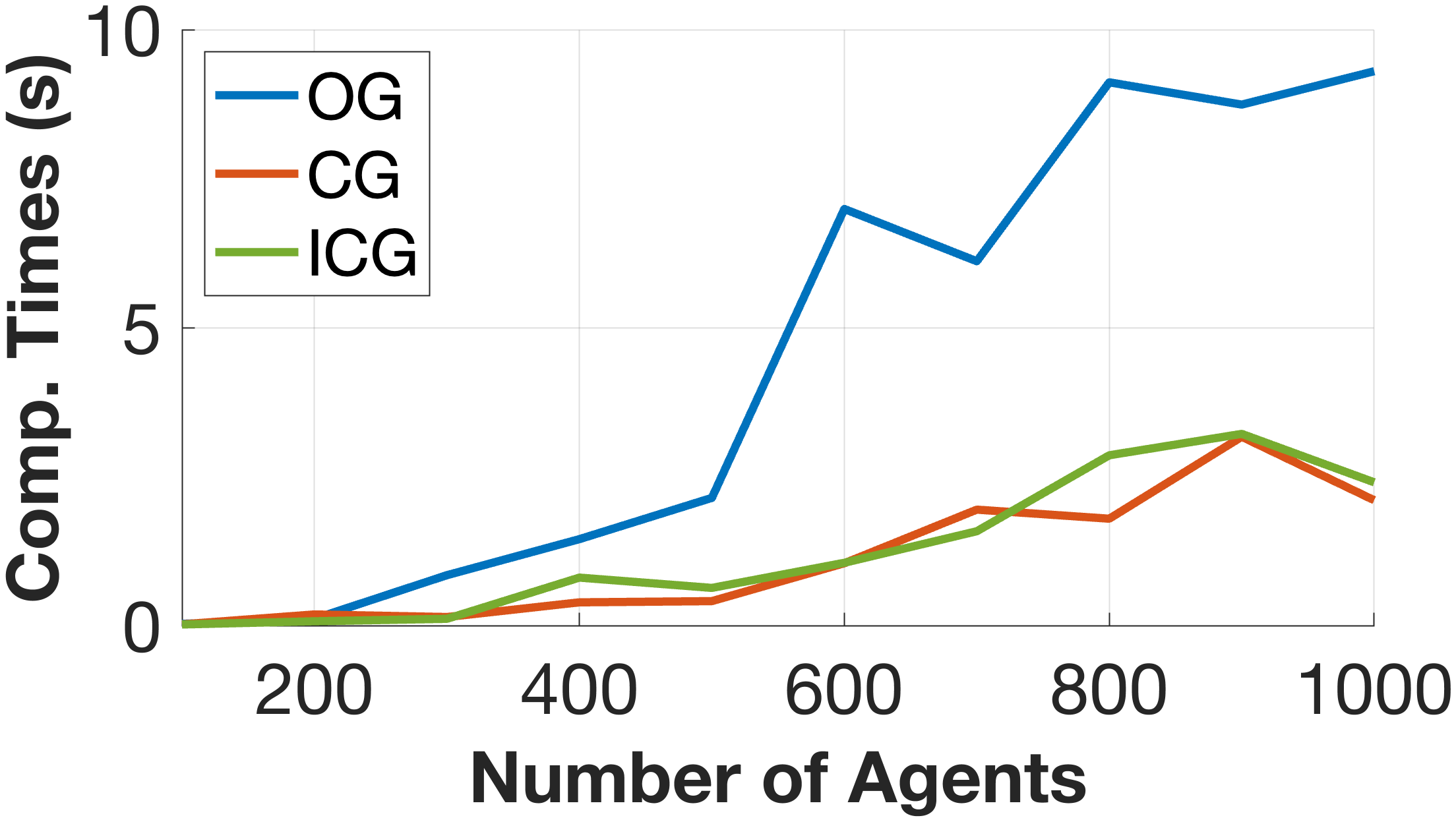}
         \caption{MAPF-LNS2 computation times}
         \label{fig:global-lns-comps}
     \end{subfigure}
     \hfill
     \begin{subfigure}{0.3\textwidth}
         \centering
         \includegraphics[width=0.9\textwidth]{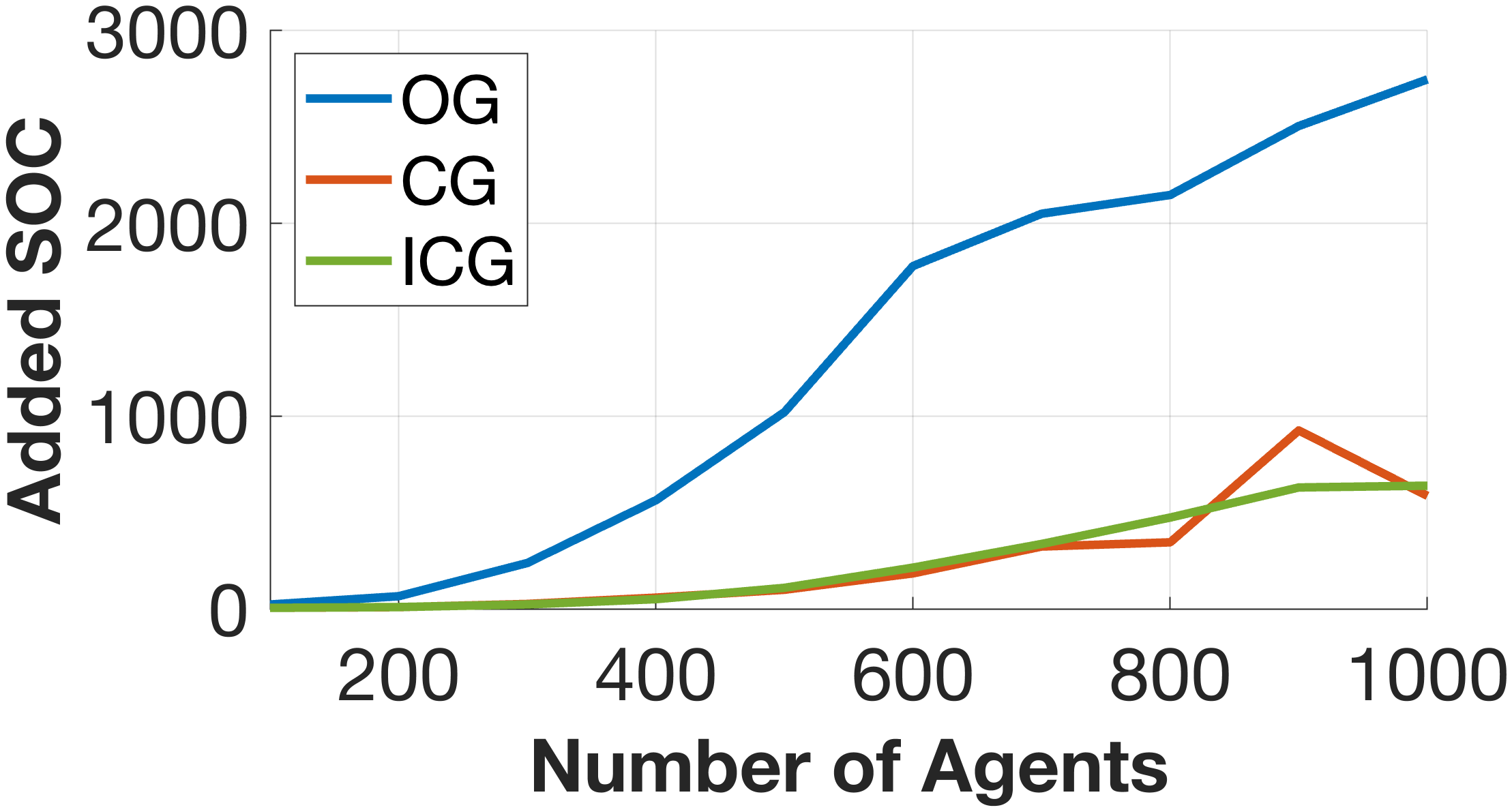}
         \caption{MAPF-LNS2 added SOC}
         \label{fig:global-lns-apl}
     \end{subfigure}
     \caption{Global averages over all tested instances. CG and ICG lines are coincident in Figs.~\ref{fig:global-cbs-succ}-\ref{fig:global-eecbs-succ}, and~\ref{fig:global-lns-succ}.}
     \label{fig:global-all}
\end{figure*}
We now provide an empirical evaluation comparing different approaches for plan repair. Our approach is as follows. We take an existing plan $P$ for some MAPF instance, and we introduce 
delays to 
it such that the plan becomes colliding. We then consider three approaches to repair the plan:
\begin{itemize}
    \item In the first approach, we simply try to find a new plan from the agents' current locations to their original goals on the \emph{original graph (OG)} using a MAPF solver.
    \item In the second (resp. third) approach, we implement our reduction to the \emph{Constrained Graph (CG)} from \cref{subsec:CG} (resp. \emph{Improved Constrained Graph (ICG)} from \cref{subsec:ICG}), and use a MAPF solver.
\end{itemize}
The approaches above are further split according to which MAPF solver we use, as we detail in~\cref{subsec:setup}. 
An important point is that we test on CBS, which typically does not scale very well without suboptimal heuristics. 

We remark that a-priori, the comparison with replanning on OG is not ``fair'', in that OG allows for shorter plans that are unavailable when only delays can be introduced. Nonetheless, we show that when only a single delay is introduced, our approach is competitive also in this sense --- we almost never output longer plans than OG (especially as the number of agents increases) with the exception of a few outliers.
Thus, our approach has both the advantage of keeping to the original plan, as discussed in \cref{sec:intro}, as well as in plan length.



In our setting, we consider environments where delays are enabled on all vertices (c.f. \cref{rmk:variants_of_ACID}). While this is not used by 
the algorithms, it does allow us to strengthen \cref{lem:quadratic_upper_bound}.
\begin{remark}[Upper bound on delays]
 \label{rmk:upper_bound_delays}
Consider a colliding plan $P$ for $n$ agents obtained by experiencing $d$ unexpected delay from a non-colliding plan. We can always repair $P$ by delaying all the remaining agents for $d$ timesteps, hence synchronizing back to the non-colliding plan. This gives an upper bound of $d(n-1)$ on the number of delays necessary to repair a delayed plan.
\end{remark}

\noindent
We can use~\cref{rmk:upper_bound_delays} as a sanity check on the optimality of solutions obtained in the experiments.
We divide our experiments to ones where we introduce a single delay, and ones where we introduce multiple delays. Conceptually, the former is a ``cleaner'' study of the effect of a delay. As we show, however, 
the latter is much more challenging to solve.

\subsection{Experimental Setup -- Single Delay}
\label{subsec:setup}
We consider nine different MAPF maps from~\citet{stern2019multi} (see  Fig.~\ref{fig:mapf-maps}).
For each map, we selected $10$ random MAPF instances and use Anytime-EECBS~\cite{Li_Ruml_Koenig_2021}
to calculate a high-quality MAPF solution $P$ given a time budget of three minutes.
If successful, we perform $10$ iterations where, in each iteration, we sample a \emph{collision-inducing} delay for a random agent $i$ at a random step $0 < k < m_i$, where $m_i$ is the agent $i$'s path length, such that the resulting plan becomes colliding at $k < t < m_i$.
We then attempt to repair this plan on OG, CG and ICG using CBS, Anytime-EECBS, and MAPF-Large Neighborhood Search 2 (MAPF-LNS2)~\cite{Li_Chen_Harabor_Stuckey_Koenig_2022} within a three minute time limit.

We emphasize that our goal is not to compare MAPF algorithms, but to compare the effect of CG and ICG against OG on different MAPF algorithms.

The number of agents was incremented from $n=100$ in steps of $100$ 
until a maximum number of agents was reached for a particular instance. 
For most instances, the maximal number of agents was $n=1000$.

We evaluated our findings using three metrics:
\begin{enumerate}[label={(\roman*)}]
    \item success rate (i.e., percentage of plans for which a solution was computed within the allotted time budget),
    \item computation time (only on successful instances), and
    \item added plan length. For CG and ICG, this amounts to the number of delays introduced.
\end{enumerate}
All evaluations were performed on AMD 4.5 GHz CPU and 64 GB of RAM.
Our implementation\footnote{\url{https://github.com/aria-systems-group/Delay-Robust-MAPF}}
was forked from the MAPF-LNS2 codebase\footnote{\url{https://github.com/Jiaoyang-Li/MAPF-LNS2}.} built in C++.
Due to space constraints, we provide a representative subset of our results here. 

We remark that 
we do not seed the MAPF solvers (e.g., MAPF-LNS2) with the colliding plan because we are only interested in the affect of CG/ICG compared to OG. Doing so is an implementation choice that does not affect our conclusions in any significant way. 
 \subsection{Computation Time}
\begin{table*}[t]
    \centering
    \begin{tabular}{cc@{\quad}SSScSSScSSS}
        \multirow{2}{*}{Row} & \multirow{2}{*}{Map} & \multicolumn{3}{c}{CBS} & & \multicolumn{3}{c}{Anytime-EECBS} & & \multicolumn{3}{c}{MAPF-LNS2} \\
        \cmidrule(r){3-5}\cmidrule(lr){7-9}\cmidrule(lr){11-13}
        & & OG & CG & ICG & & OG & CG & ICG & & OG & CG & ICG \\
        \midrule
1 & Fig.~\ref{fig:Berlin_1_256} & 12.5 & \textbf{4.8}  & \textbf{4.6} & & 10.6 & 12.7  & \textbf{6.8} & & 7.0 & \textbf{0.8}  & \textbf{0.5} \\
2 & Fig.~\ref{fig:Boston_0_256} & 6.5 & \textbf{4.9}  & \textbf{4.7}  & & 13.3 & 14.3  & \textbf{8.3}  & & 13.4 & \textbf{1.3}   & \textbf{2.6} \\
3 & Fig.~\ref{fig:den520d} & \text{---} & \textbf{9.7}   & \textbf{15.5}  & & 18.4 & 20.6  & 24.0 & & 44.2 & \textbf{3.1}   & \textbf{4.0}  \\
4 & Fig.~\ref{fig:empty-32-32} & \text{---} & \textbf{5.7}   & \textbf{4.8}  & & 3.0 & 3.6  & \textbf{1.9}  & & 23.3 & \textbf{4.5}   & \textbf{9.6}  \\
5 & Fig.~\ref{fig:Paris_1_256} & 10.3 & \textbf{4.5 }  & \textbf{4.1}  & & 8.4 & 11.1  & \textbf{5.0}  & & 4.3 & \textbf{0.6}   & \textbf{0.7}  \\
6 & Fig.~\ref{fig:random-64-64-10} & 65.6 & \textbf{4.1}   & \textbf{3.5}  & & 0.8 & 2.3  & 2.2 & & 17.7 & \textbf{4.4}   & \textbf{13.2}  \\
7 & Fig.~\ref{fig:w_woundedcoast} & \text{---} & \textbf{23.0}   & \textbf{22.7}  & & 4.4 & 20.6  & \textbf{3.6}  & & 1.4 & 36.3  & 31.3 \\
8 & Fig.~\ref{fig:warehouse-10-20-10-2-2} & \text{---} & \textbf{6.6}  & \textbf{7.4}  & & 16.7 & \textbf{16.5}   & 25.2 & & 35.4 & \textbf{5.0}   & \textbf{2.6}  \\
9 & Fig.~\ref{fig:warehouse-20-40-10-2-2} & 25.8 & \textbf{7.2}   & \textbf{7.3}  & & 6.4 & 7.8  & \textbf{4.5}  & & 1.2 & \textbf{0.4}   & \textbf{0.4}  \\
    \end{tabular}
    \caption{Computation time means ($s$) of the initial solutions for all maps on scenes with the most agents.}
    \label{tab:comps_table}
\end{table*}
\begin{table*}[t]
    \centering
    \begin{tabular}{cc@{\quad}SSScSSScSSS}
        \multirow{2}{*}{Row} & \multirow{2}{*}{Map} & \multicolumn{3}{c}{CBS} & & \multicolumn{3}{c}{Anytime-EECBS} & & \multicolumn{3}{c}{MAPF-LNS2} \\
        \cmidrule(r){3-5}\cmidrule(lr){7-9}\cmidrule(lr){11-13}
        & & OG & CG & ICG & & OG & CG & ICG & & OG & CG & ICG \\
        \midrule
1 & Fig.~\ref{fig:Berlin_1_256} & 4.2 & \textbf{90.3} & \textbf{90.3}  & & 100.0 & 98.6 & 98.6 & & 100.0 & \textbf{100.0}  & \textbf{ 100.0} \\
2 & Fig.~\ref{fig:Boston_0_256} & 4.7 & \textbf{81.3}  & \textbf{81.3}  & & 100.0 & 98.4 & \textbf{100.0}  & & 100.0 & 96.9 & 98.4 \\
3 & Fig.~\ref{fig:den520d} & 0.0 & \textbf{61.2}  & \textbf{63.3}  & & 89.8 & 67.3 & 73.5 & & 63.3 & \textbf{89.8}  & \textbf{93.9}  \\
4 & Fig.~\ref{fig:empty-32-32} & 0.0 & \textbf{46.3}  & \textbf{46.3}  & & 7.3 & \textbf{26.8}  & \textbf{26.8}  & & 19.5 & 17.1 & \textbf{19.5}  \\
5 & Fig.~\ref{fig:Paris_1_256} & 6.5 & \textbf{90.9}  & \textbf{90.9}  & & 98.7 & \textbf{98.7}  & \textbf{98.7}  & & 100.0 & \textbf{100.0}  & \textbf{100.0}  \\
6 & Fig.~\ref{fig:random-64-64-10} & 2.4 & \textbf{61.0}  & \textbf{61.0}  & & 17.1 & \textbf{31.7}  & \textbf{34.1}  & & 22.0 & \textbf{41.5}  & \textbf{51.2}  \\
7 & Fig.~\ref{fig:w_woundedcoast} & 0.0 & \textbf{33.3}  & \textbf{33.3}  & & 12.5 & \textbf{16.7}  & \textbf{12.5}  & & 16.7 & \textbf{29.2}  & \textbf{29.2}  \\
8 & Fig.~\ref{fig:warehouse-10-20-10-2-2} & 0.0 & \textbf{83.6}  & \textbf{83.6}  & & 67.2 & 49.2 & 52.5 & & 44.3 & \textbf{93.4}  & \textbf{91.8}  \\
9 & Fig.~\ref{fig:warehouse-20-40-10-2-2} & 10.0 & \textbf{97.1}  & \textbf{97.1}  & & 97.1 & \textbf{100.0}  & \textbf{100.0}  & & 98.6 & \textbf{100.0}  & \textbf{100.0}  \\
    \end{tabular}
    \caption{Success rate means ($\%$) of the initial solutions for all maps on scenes with the most agents.}
    \label{tab:success_rate_table}
\end{table*}

\begin{table*}[th!]
    \centering
    \begin{tabular}{cc@{\quad}S[group-separator = {,},group-four-digits = true,table-format = 3.0]S[group-separator = {,},group-four-digits = true,table-format = 2.0]S[group-separator = {,},group-four-digits = true,table-format = 2.0]cS[group-separator = {,},group-four-digits = true,table-format = 5.0]S[group-separator = {,},group-four-digits = true,table-format = 4.0]S[group-separator = {,},group-four-digits = true,table-format = 4.0]cS[group-separator = {,},group-four-digits = true,table-format = 4.0]S[group-separator = {,},group-four-digits = true,table-format = 5.0]S[group-separator = {,},group-four-digits = true,table-format = 5.0]}
        \multirow{2}{*}{Row} & \multirow{2}{*}{Map} & \multicolumn{3}{c}{CBS} & & \multicolumn{3}{c}{Anytime-EECBS} & & \multicolumn{3}{c}{MAPF-LNS2} \\
        \cmidrule(r){3-5}\cmidrule(lr){7-9}\cmidrule(lr){11-13}
        & & OG & CG & ICG & & OG & CG & ICG & & OG & CG & ICG \\
        \midrule
1 & Fig.~\ref{fig:Berlin_1_256} & -16 & 4  & 4 & & 1727 & \textbf{1037}   & \textbf{ 1036} & & 2977 & \textbf{171}   & \textbf{199}  \\
2 & Fig.~\ref{fig:Boston_0_256} & -16 & 3  & 3 & & 4813 & \textbf{2803}   & \textbf{2816}  & & 6428 & \textbf{995}  & \textbf{983}  \\
3 & Fig.~\ref{fig:den520d} & \text{---} & \textbf{4}   & \textbf{4}  & & 10305 & \textbf{5847}   & \textbf{6141}  & & 9297 & \textbf{1019}   & \textbf{1036}  \\
4 & Fig.~\ref{fig:empty-32-32} & \text{---} & \textbf{11}   & \textbf{11}  & & 457 & \textbf{262}   & \textbf{340}  & & 420 & 480  & \textbf{375}  \\
5 & Fig.~\ref{fig:Paris_1_256} & -8 & 3  & 3 & & 1110 & \textbf{694}   & \textbf{695}  & & 2114 & \textbf{141}   & \textbf{144}  \\
6 & Fig.~\ref{fig:random-64-64-10} & -37 & 5  & 5 & & 607 & \textbf{555}   & 661 & & 1340 & \textbf{726}   & \textbf{1053}  \\
7 & Fig.~\ref{fig:w_woundedcoast} & \text{---} & \textbf{5}   & \textbf{5}  & & 1657 & 5926  & \textbf{1026}  & & 415 & 12950  & 14760 \\
8 & Fig.~\ref{fig:warehouse-10-20-10-2-2} & \text{---} & \textbf{5}   & \textbf{5}  & & 5245 & \textbf{1251}   & \textbf{1363}  & & 4074 & \textbf{540}   & \textbf{487}  \\
9 & Fig.~\ref{fig:warehouse-20-40-10-2-2} & -11 &  3  &  3 & & 506 & \textbf{293}   & \textbf{293}  & & 767 & \textbf{73}   & \textbf{75}  \\
    \end{tabular}
    \caption{Added SOC means of the initial solutions for all maps on scenes with the most agents.}
    \label{tab:soc-means}
\end{table*}
\begin{table*}[th!]
    \footnotesize
    \centering
    \begin{tabular}{cc@{\quad}ccccccccccc}
        Delay & \multirow{2}{*}{Statistic} & \multicolumn{3}{c}{CBS} & & \multicolumn{3}{c}{Anytime-EECBS} & & \multicolumn{3}{c}{MAPF-LNS2} \\
        \cmidrule(r){3-5}\cmidrule(lr){7-9}\cmidrule(lr){11-13}
        Introduction & & OG & CG & ICG & & OG & CG & ICG & & OG & CG & ICG \\
        \midrule
\multirow{3}{*}{10} & Succ. Rate (\%) & 1 & \textbf{32} & \textbf{32} &  & 99 & 98 & 98 &  & 32 & \textbf{95} & \textbf{100} \\
& Comp. Times (s) & \text{---} & \textbf{5.3} & \textbf{5.3} &  & 1.6 & 2.1 & 1.7 &  & 5.4 & \textbf{3.7} & \textbf{4.5} \\
& Added SOC & \text{---} & \textbf{7} & \textbf{7} &  & 587 (168) & \textbf{346} (\textbf{80}) & \textbf{346} (\textbf{79}) &  & 230 (66) & \textbf{136} (99) & \textbf{140} (102) \\
        \midrule
\multirow{3}{*}{50} & Succ. Rate (\%) & 0 & 0 & 0 &  & 100 & \textbf{100} & \textbf{100} &  & 31 & \textbf{100} & \textbf{100} \\
& Comp. Times (s) & \text{---} & \text{---} & \text{---} &  & 1.8 & 2.8 & 2.1 &  & 7.1 & \textbf{5.2} & \textbf{5.5} \\
& Added SOC & \text{---} & \text{---} & \text{---} &  & 660 (318) & 758 (\textbf{296}) & 757 (\textbf{314}) &  & 336 (138) & 556 (370) & 567 (379) \\
    \end{tabular}
    \caption{Experimental results for $10$ and $50$ simultaneous delays. Anytime algorithms  report two values for Added SOC: the initial solution and the best solution (in parentheses). Computation times are provided for the initial solutions.}
    \label{tab:multi-delay-stats}
\end{table*}
The average computation times of every MAPF solver over all of the instances are shown in Figs.~\ref{fig:global-cbs-comps},~\ref{fig:global-eecbs-comps}, and~\ref{fig:global-lns-comps}. The three lines correspond to OG (blue), CG (orange), and ICG (green). The most obvious computation time improvements are seen in Fig.~\ref{fig:global-cbs-comps} which show that not only can CBS on CG and ICG scale to $1000$ agents, but it also 
produced very low computation times compared to OG. Similarly, MAPF-LNS2 on CG and ICG performed faster than on OG (see Fig.~\ref{fig:global-lns-comps}). Anytime-EECBS on ICG improved its computation times over both CG and OG, especially for large number of agents (see Fig.~\ref{fig:global-eecbs-comps}). 

The computation times for all three MAPF algorithms on the
most difficult scenarios are
shown in Table~\ref{tab:comps_table}. Combinations that performed at least as good as OG are \textbf{bolded}. 

CBS on CG and ICG consistently outperformed
OG (Table~\ref{tab:comps_table}, rows 1-9). 
Interestingly, Anytime-EECBS is generally faster on OG than CG (Table~\ref{tab:comps_table}, all rows except 8) but is faster on ICG than both OG and CG (Table~\ref{tab:comps_table}, all rows except 3 and 7). MAPF-LNS2 shows obvious computation-time improvements of up to $14\times$ than when using OG (Table~\ref{tab:comps_table}, all rows except row 7).

\subsection{Success Rate}
The average success rates of every MAPF solver over all the instances are shown in Figs.~\ref{fig:global-cbs-succ},~\ref{fig:global-eecbs-succ}, and~\ref{fig:global-lns-succ}. The results show that CG and ICG perform better than OG across all values of $n$ for both CBS and MAPF-LNS2 while performing equally as well for Anytime-EECBS. CBS improved the most, which succeeded over $80\%$ of the time for all $n$ with CG and ICG but only scaled to~$400$ agents on OG.

The success rates for all three MAPF algorithms on the most difficult (largest number of agents $n$) scenarios are shown in Table~\ref{tab:success_rate_table}. The entries where CG or ICG performed at least as well as OG are \textbf{bolded}.

Observe that using CG and ICG dramatically improved ($9\times$, at least) the success rate of CBS on all tested maps (rows 1-9 of Table~\ref{tab:success_rate_table}). In addition, the optimal CBS algorithm becomes a viable candidate for solving instances of ACID up to $1000$ agents.
Anytime-EECBS provided similar success rates for all three graphs on most examples (rows 1, 2, 5, 7 and 9 of Table~\ref{tab:success_rate_table}). There are however, scenarios where Anytime-EECBS both improved the success rate (rows 4 and 6 of Table~\ref{tab:success_rate_table}) and hindered it (rows 3, and 8 of Table~\ref{tab:success_rate_table}).
MAPF-LNS2 produced similar success rates for all three graphs on about half of the tested instances (rows 1, 2, 4, 5, and 9) but did show significant improvements on the other half (rows 3, and 6-8 of Table~\ref{tab:success_rate_table}). 

\subsection{Added SOC}
The average added SOC of every MAPF solver over all the maps are shown in Figs.~\ref{fig:global-cbs-apl},~\ref{fig:global-eecbs-apl}, and~\ref{fig:global-lns-apl}. The results show that both Anytime-EECBS and MAPF-LNS2 generally provided shorter solutions when using CG and ICG compared to OG (see Figs.~\ref{fig:global-eecbs-apl} and~\ref{fig:global-lns-apl}). Replanning with CBS on CG and ICG generally results in very small SOC additions (about $4\ll n$ as per Remark~\ref{rmk:upper_bound_delays}) compared to Anytime-EECBS and MAPF-LNS2 (see Fig.~\ref{fig:global-cbs-apl}).

The added SOC for all three MAPF algorithms on the most difficult (largest number of agents $n$) scenarios are shown in Table~\ref{tab:soc-means}. The entries where CG or ICG performed at least as well as OG are shown in \textbf{bold}.

In terms of added SOC, we see that in the (\emph{very}) rare occasion that CBS succeeds on OG, it can improve the original plan (rows 1-2, 5-6, and 9 of Table~\ref{tab:soc-means}). Meanwhile, CBS on  CG and ICG generally repair plans with $1000$ agents with at most $11$ additional delays (all rows of Table~\ref{tab:soc-means}). Anytime-EECBS produces a larger number of delays compared to CBS and MAPF-LNS2 but using CG and ICG performs better than OG, in general (see rows 1-9 of Table~\ref{tab:soc-means}). MAPF-LNS2 on CG and ICG generally improved optimality compared to OG (all rows except 7 of Table~\ref{tab:soc-means}).

Overall, planning on CG and ICG improved all three algorithms in different aspects. CBS on CG and ICG scaled to $1000$ agents while minimizing added plan length. CG and ICG also enabled Anytime-EECBS to produce more optimal solutions while occasionally improving success rate. MAPF-LNS2 on CG and ICG greatly improved the added plan length and success rates compared to OG while also improving computation times. 

Note that the performance differences on CG and ICG are typically very small, if any. This occurs because as the space becomes congested, most vertices become intersections, and hence CG and ICG become almost identical.
\subsection{Experimental Setup – Multiple Delays}
Our setup for introducing multiple delays is identical to that of Section~\ref{subsec:setup} with some key exceptions: we introduced multiple simultaneous delays, we  tested on $3$ maps (Figs.~\ref{fig:Paris_1_256},~\ref{fig:random-64-64-10}, and~\ref{fig:warehouse-10-20-10-2-2}), we only considered instances with $600$ agents, our timeout was \emph{one} minute, and we considered the anytime properties of Anytime-EECBS and MAPF-LNS2. Table~\ref{tab:multi-delay-stats} report our results for $10$ and $50$ delay introductions. 

Roughly summarizing our findings, we see that CBS fails almost entirely in the presence of multiple delays. This fact is interesting, as it identifies a class of graphs for which CBS struggles that is not caused due to the number of agents or the branching degree (as CBS works well in the presence of one delay with even more agents).

We see that Anytime-EECBS performs roughly equally well on OG, CG and ICG, but typically elongates the plan more on OG than in CG and ICG (despite having potentially shorter plans!). MAPF-LNS2 performs poorly on OG, but well on CG and ICG, introducing less delays than Anytime-EECBS, when successful.

We conclude that for multiple delays, heuristic algorithms are preferable to the optimal CBS, as expected, but that CG and ICG do assist 
in minimizing the number of delay. 
\section{Conclusion}
We address the issue of repairing MAPF plans after encountering unexpected delays. We introduce the ACID problem, and prove it is \NP-Complete. We adapt ACID into an MAPF problem using two novel graph formulations, CG and ICG, which confine the graph to the original paths. 
We empirically show that the CG and ICG improve MAPF algorithms' ability to repair plans compared to traditional MAPF.

\section*{Acknowledgements}
This research was supported in by the Israeli Ministry of Science \& Technology grants No. 3-16079 and 3-17385, the United States-Israel Binational Science Foundation (BSF) grants no. 2019703 and 2021643, the Israel Science Foundation (grant nos.~1736/19 and~2261/23), by NSF/US-Israel-BSF (grant no.~2019754), by the Blavatnik Computer Science Research Fund, the Israeli Smart Transportation Research Center (ISTRC), Israel Science Foundation (grant No. 989/22), and the University of Colorado Boulder. 
\bibliography{main}

\begin{thebibliography}{30}
\providecommand{\natexlab}[1]{#1}

\bibitem[{Abrahamsen et~al.(2023)Abrahamsen, Geft, Halperin, and
  Ugav}]{abrahamsen2023coordination}
Abrahamsen, M.; Geft, T.; Halperin, D.; and Ugav, B. 2023.
\newblock Coordination of Multiple Robots along Given Paths with Bounded
  Junction Complexity.
\newblock In \emph{{AAMAS}}, 932--940.

\bibitem[{Atzmon et~al.(2020{\natexlab{a}})Atzmon, Stern, Felner, Sturtevant,
  and Koenig}]{AtzmonSFSK20}
Atzmon, D.; Stern, R.; Felner, A.; Sturtevant, N.~R.; and Koenig, S.
  2020{\natexlab{a}}.
\newblock Probabilistic Robust Multi-Agent Path Finding.
\newblock In \emph{{ICAPS}}, 29--37.

\bibitem[{Atzmon et~al.(2020{\natexlab{b}})Atzmon, Stern, Felner, Wagner,
  Bart{\'{a}}k, and Zhou}]{AtzmonSFWBZ20}
Atzmon, D.; Stern, R.; Felner, A.; Wagner, G.; Bart{\'{a}}k, R.; and Zhou, N.
  2020{\natexlab{b}}.
\newblock Robust Multi-Agent Path Finding and Executing.
\newblock \emph{{JAIR}}, 67: 549--579.

\bibitem[{Ausiello et~al.(1999)Ausiello, Marchetti{-}Spaccamela, Crescenzi,
  Gambosi, Protasi, and Kann}]{APX-background}
Ausiello, G.; Marchetti{-}Spaccamela, A.; Crescenzi, P.; Gambosi, G.; Protasi,
  M.; and Kann, V. 1999.
\newblock \emph{Complexity and approximation: combinatorial optimization
  problems and their approximability properties}.
\newblock Springer.

\bibitem[{Banfi, Basilico, and Amigoni(2017)}]{Banfi2017}
Banfi, J.; Basilico, N.; and Amigoni, F. 2017.
\newblock {Intractability of Time-Optimal Multirobot Path Planning on 2D Grid
  Graphs with Holes}.
\newblock \emph{{Robot. Autom. Lett.}}, 2: 1941--1947.

\bibitem[{Bart{\'a}k, {\v{S}}vancara, and Vlk(2018)}]{bartak2018scheduling}
Bart{\'a}k, R.; {\v{S}}vancara, J.~{\'\i}.; and Vlk, M. 2018.
\newblock A scheduling-based approach to multi-agent path finding with weighted
  and capacitated arcs.
\newblock In \emph{{AAMAS}}, 748--756.

\bibitem[{Belov et~al.(2020)Belov, Du, de~la Banda, Harabor, Koenig, and
  Wei}]{BelovDBHKW20}
Belov, G.; Du, W.; de~la Banda, M.~G.; Harabor, D.; Koenig, S.; and Wei, X.
  2020.
\newblock From Multi-Agent Pathfinding to 3D Pipe Routing.
\newblock In \emph{{SOCS}}, 11--19.

\bibitem[{Berndt et~al.(2020)Berndt, Duijkeren, Palmieri, and
  Keviczky}]{berndt2020feedback}
Berndt, A.; Duijkeren, N.~V.; Palmieri, L.; and Keviczky, T. 2020.
\newblock A Feedback Scheme to Reorder a Multi-Agent Execution Schedule by
  Persistently Optimizing a Switchable Action Dependency Graph.
\newblock arXiv:2010.05254.

\bibitem[{DeHaan and Friggstad(2023)}]{SumColoring-APX}
DeHaan, I.; and Friggstad, Z. 2023.
\newblock Approximate Minimum Sum Colorings and Maximum k-Colorable Subgraphs
  of Chordal Graphs.
\newblock In \emph{{WADS}}, volume 14079 of \emph{Lecture Notes in Computer
  Science}, 326--339. Springer.

\bibitem[{Geft(2023)}]{DBLP:conf/socs/Geft23}
Geft, T. 2023.
\newblock Fine-Grained Complexity Analysis of Multi-Agent Path Finding on 2D
  Grids.
\newblock In \emph{{SOCS}}, 20--28. {AAAI} Press.

\bibitem[{H{\"{o}}nig et~al.(2016)H{\"{o}}nig, Kumar, Cohen, Ma, Xu, Ayanian,
  and Koenig}]{Hoenig_Kumar_Cohen_Ma_Xu_Ayanian_Koenig_2016}
H{\"{o}}nig, W.; Kumar, T. K.~S.; Cohen, L.; Ma, H.; Xu, H.; Ayanian, N.; and
  Koenig, S. 2016.
\newblock Multi-Agent Path Finding with Kinematic Constraints.
\newblock In \emph{{ICAPS}}, 477--485.

\bibitem[{Komenda and Nov{\'a}k(2011)}]{komenda2011multi}
Komenda, A.; and Nov{\'a}k, P. 2011.
\newblock Multi-agent plan repairing.
\newblock In \emph{Decision Making in Partially Observable, Uncertain Worlds:
  Exploring Insights from Multiple Communities, Proceedings of IJCAI 2011
  Workshop}, 1--6.

\bibitem[{Komenda, Novák, and Pěchouček(2014)}]{KOMENDA201476}
Komenda, A.; Novák, P.; and Pěchouček, M. 2014.
\newblock Domain-independent multi-agent plan repair.
\newblock \emph{J. Netw Comput Appl}, 37: 76--88.

\bibitem[{Kubicka and Schwenk(1989)}]{SumColoring}
Kubicka, E.~M.; and Schwenk, A.~J. 1989.
\newblock An Introduction to Chromatic Sums.
\newblock In \emph{{ACM} Conf. on CS}, 39--45.

\bibitem[{Li et~al.(2022)Li, Chen, Harabor, Stuckey, and
  Koenig}]{Li_Chen_Harabor_Stuckey_Koenig_2022}
Li, J.; Chen, Z.; Harabor, D.; Stuckey, P.~J.; and Koenig, S. 2022.
\newblock {MAPF-LNS2:} Fast Repairing for Multi-Agent Path Finding via Large
  Neighborhood Search.
\newblock In \emph{{AAAI}}, 10256--10265.

\bibitem[{Li et~al.(2021)Li, Chen, Zheng, Chan, Harabor, Stuckey, Ma, and
  Koenig}]{LCZCHS0K21}
Li, J.; Chen, Z.; Zheng, Y.; Chan, S.; Harabor, D.; Stuckey, P.~J.; Ma, H.; and
  Koenig, S. 2021.
\newblock Scalable Rail Planning and Replanning: Winning the 2020 Flatland
  Challenge.
\newblock In \emph{{ICAPS}}, 477--485.

\bibitem[{Li, Ruml, and Koenig(2021)}]{Li_Ruml_Koenig_2021}
Li, J.; Ruml, W.; and Koenig, S. 2021.
\newblock {EECBS:} {A} Bounded-Suboptimal Search for Multi-Agent Path Finding.
\newblock In \emph{{AAAI}}, 12353--12362.

\bibitem[{Ma et~al.(2019)Ma, Harabor, Stuckey, Li, and
  Koenig}]{ma2019searching}
Ma, H.; Harabor, D.; Stuckey, P.~J.; Li, J.; and Koenig, S. 2019.
\newblock Searching with Consistent Prioritization for Multi-Agent Path
  Finding.
\newblock In \emph{{AAAI}}, 7643--7650.

\bibitem[{Ma, Kumar, and Koenig(2017)}]{Ma_Kumar_Koenig_2017}
Ma, H.; Kumar, T. K.~S.; and Koenig, S. 2017.
\newblock Multi-Agent Path Finding with Delay Probabilities.
\newblock In \emph{{AAAI}}, 3605--3612.

\bibitem[{Nebel(2020)}]{nebel2020computational}
Nebel, B. 2020.
\newblock On the Computational Complexity of Multi-Agent Pathfinding on
  Directed Graphs.
\newblock In \emph{{ICAPS}}, 212--216.

\bibitem[{Nebel and Koehler(1995)}]{nebel1995plan}
Nebel, B.; and Koehler, J. 1995.
\newblock Plan reuse versus plan generation: A theoretical and empirical
  analysis.
\newblock \emph{Art. Int.}, 76(1-2): 427--454.

\bibitem[{Okumura(2023)}]{Okumura_2023}
Okumura, K. 2023.
\newblock {LaCAM}: Search-Based Algorithm for Quick Multi-Agent Pathfinding.
\newblock In \emph{{AAAI}}, 11655--11662.

\bibitem[{Ramaithitima et~al.(2016)Ramaithitima, Whitzer, Bhattacharya, and
  Kumar}]{7395308}
Ramaithitima, R.; Whitzer, M.; Bhattacharya, S.; and Kumar, V. 2016.
\newblock Automated Creation of Topological Maps in Unknown Environments Using
  a Swarm of Resource-Constrained Robots.
\newblock \emph{{Robot. Autom. Lett.}}, 1(2): 746--753.

\bibitem[{Salzman and Stern(2020)}]{SalzmanS20}
Salzman, O.; and Stern, R. 2020.
\newblock Research Challenges and Opportunities in Multi-Agent Path Finding and
  Multi-Agent Pickup and Delivery Problems.
\newblock In \emph{{AAMAS}}, 1711--1715.

\bibitem[{Sharon et~al.(2015)Sharon, Stern, Felner, and
  Sturtevant}]{sharon2015conflict}
Sharon, G.; Stern, R.; Felner, A.; and Sturtevant, N.~R. 2015.
\newblock Conflict-based search for optimal multi-agent pathfinding.
\newblock \emph{Art. Int.}, 219: 40--66.

\bibitem[{Stern et~al.(2019)Stern, Sturtevant, Felner, Koenig, Ma, Walker, Li,
  Atzmon, Cohen, Kumar, Bart{\'{a}}k, and Boyarski}]{stern2019multi}
Stern, R.; Sturtevant, N.~R.; Felner, A.; Koenig, S.; Ma, H.; Walker, T.; Li,
  J.; Atzmon, D.; Cohen, L.; Kumar, S.; Bart{\'{a}}k, R.; and Boyarski, E.
  2019.
\newblock Multi-Agent Pathfinding: Definitions, Variants, and Benchmarks.
\newblock In \emph{{SOCS}}, 151--159.

\bibitem[{Svancara et~al.(2023)Svancara, Tignon, Bart{\'{a}}k, Schaub, Wanko,
  and Kaminski}]{vsvancara2023multi}
Svancara, J.; Tignon, E.; Bart{\'{a}}k, R.; Schaub, T.; Wanko, P.; and
  Kaminski, R. 2023.
\newblock Multi-Agent Pathfinding with Predefined Paths: To Wait, or Not to
  Wait, That Is the Question [Extended Abstract].
\newblock In \emph{{SOCS}}, 185--186.

\bibitem[{Tonola et~al.(2023)Tonola, Faroni, Beschi, and Pedrocchi}]{10013661}
Tonola, C.; Faroni, M.; Beschi, M.; and Pedrocchi, N. 2023.
\newblock Anytime Informed Multi-Path Replanning Strategy for Complex
  Environments.
\newblock \emph{IEEE Access}, 11: 4105--4116.

\bibitem[{Wurman, D’Andrea, and Mountz(2008)}]{wurman2008cooperative}
Wurman, P.~R.; D’Andrea, R.; and Mountz, M. 2008.
\newblock Coordinating Hundreds of Cooperative, Autonomous Vehicles in
  Warehouses.
\newblock \emph{AI Magazine}, 29(1): 9.

\bibitem[{Yu(2016)}]{Yu2016}
Yu, J. 2016.
\newblock {Intractability of optimal multirobot path planning on planar
  graphs}.
\newblock \emph{{Robot. Autom. Lett.}}, 1: 33--40.

\end{thebibliography}
\clearpage

\section{Technical Appendix}
\label{sec:appendix}
We now present the complete proofs of Theorems~\ref{thm:ACID_NP_complete} and~\ref{thm:ACID_equiv_CG_equiv_ICG}.
\subsection{Proof of Theorem~\ref{thm:ACID_NP_complete}}
\begin{proof}
We present the complete proof of the \NP-hardness of ACID. Formally, the reduction proceeds as follows. 
Consider a graph $G=\tup{V,E}$ and a coloring sum $C\in \bbN$, where for convenience we denote $V=\{1,\ldots, n\}$ and $E=\{e_1,\ldots, e_m\}$. We construct an instance of ACID as follows. 
The agents are $A=\{v_i\mid i\in V\}$.
The underlying graph is $H=\tup{V',E'}$ where $V'=V \cup ([n] \times [m] \times [C+1])$ and $[q]=\{1,2,\ldots, q\}$.
Intuitively, the graph $H$ is composed of $C+1$ identical blocks, i.e., each $([n] \times [m] \times \{\ell\})$ is a block denoted by $B_{\ell}$, and $n$ additional vertices where the agents start.
We define the edges in each block by defining the plan $P$ of paths for each agent.

Each agent's path within a given block comprises its own vertices (not appearing on other paths) except for shared vertices representing edges of $G$.
That is, for each edge $e_r=\{i,j\}$ in $G$, the $r$-th vertex of agent $i$ and agent $j$ in the block are the same vertex.
The $C+1$ blocks are joined by edges through a natural concatenation of the agents' paths.
That is, the last vertex of each path $\pi_i$ in $B_{\ell}$ connects to the first vertex of $\pi_i$ in the next block $B_{\ell+1}$ via an edge.
Also, the starting vertex of each agent leads to the agent's first block in $B_1$ via an edge.

We output an ACID instance with a budget of $C$.
We claim that we can identify any plan $\{\pi'_1,\ldots,\pi'_n\}$ obtained from $P$ using delays with a coloring for $G$ by having every vertex's color be the delay of the corresponding agent.

For the first direction, assume $\chi$ is a coloring of $G$ of sum at most $C$, we construct a non-colliding plan $P'$ where each agent $v_i$ waits $\chi(i)$ time steps at its start vertex, after which it proceeds without further delays to its target.
We verify that the resulting plan $P'$ is non-colliding.
Let $v_i$ and $v_j$ be two agents who share a vertex on their paths. This means that $\{i,j\}$ is an edge of $G$, and hence $v_i$ and $v_j$ get delayed by a different number of time steps initially since $\chi(i) \neq \chi(j)$.
Furthermore, any shared vertex along the paths of $v_i$ and $v_j$ appears the same number of steps (vertices) from their respective start positions. Consequently, $v_i$ and $v_j$ do not collide, as they are delayed by a different number of time steps.

Conversely, consider a non-colliding plan $P'$ obtained from $P$ by introducing at most $C$ delays, then by the pigeonhole principle, there is block $B_{\ell}$ in which no agent is delayed (as there are $C+1$ blocks).
We notice that we may further assume that no agent is delayed in each subsequent block $B_{\ell'}$ with $\ell' > \ell$.
Indeed, for any pair of agents $v_i, v_j$, if no further delays are introduced, then their paths through block $B_{\ell+1}$ are identical to their path through $B_{\ell}$, and are hence also not colliding. 
This argument applies to all agent pairs and all subsequent blocks, so we may  assume that the delay incurred by each agent up to $B_{\ell}$ equals the total delay for that agent in $P'$.
Consequently, for each edge $\{i, j\}$ the agents $v_i, v_j$ are assigned different delays in $P'$, and so they get a different coloring. 
Thus, the induced coloring $\chi$ satisfies $\chi(i)\neq \chi(j)$ and so we conclude that $G$ is colorable with a sum of at most $C$.

To show the hardness of approximation we use the same reduction except that the number of blocks is taken to be $U+1$, where $U$ is the upper bound given by~\Cref{lem:quadratic_upper_bound} (since we are now working with the optimization version of MSC, and so $C$ is not given).
Since we identify a solution to ACID with a solution of the same cost to MSC, we obtain a strict approximation preserving reduction, i.e., any $\alpha$ approximate solution to ACID corresponds to an $\alpha$ approximate solution to MSC. Hence, the \APX-hardness of MSC implies the same for ACID.
\end{proof}

\begin{remark}[]
    Note that using repeated blocks is crucial for the second direction of the proof.
    If we had only a single block, we would not be able to conclude that $v_i$ and $v_j$ have a different \emph{total} delay only because they do not collide. While one of the agents must indeed arrive later to the shared vertex of their paths, that agent could incur further delays later on.
\end{remark}

\subsection{Proof of Theorem~\ref{thm:ACID_equiv_CG_equiv_ICG}}
\begin{proof}
Clearly (3) implies (2), since the ICG is a subgraph of the CG. 
We show that (2) implies (1). Consider a solution~$P'$ of CG with $\|P'\|\le \|P\|+D$. Since agents are restricted to traverse their paths in $P$ or stay in prolongable vertices, we have that $P'$ is obtained from $P$ by delaying some agents for at most $D$ steps. Since $P'$ is non-colliding, it is a solution for ACID with budget at most $D$.

Finally, we show (1) implies (3). Assume the ACID instance is solvable with budget $D$. Then, there exists a non-colliding plan $P'$ with $\|P'\|\le \|P\|+D$ that is obtained from $P$ by introducing at most $D$ delays. If $P'$ describes valid paths in the ICG, we are done. Otherwise, there exists a prolongable vertex $v$ in $P$ for agent $i$ that was delayed, but whose self-edge is not in the ICG. Assume $v$ is vertex~$\pi^i_j$. 
We first assume that $j\le \max I_i$. Indeed, if $j$ occurs after the maximal intersecting vertex of $\pi_i$, then if we remove the delays on it the resulting plan is still non-colliding (since there are no future path intersections), and has a lower budget. 

Thus, let $s,t\in I_i\cup \{0\}$ be such that $s<j\le t$ (and $s$ is maximal and $t$ minimal with this property). Since $j$ is prolongable, there exists $s<j'\le t$ that is prolongable in the ICG. We obtain a plan $P''$ by replacing the delays on $\pi^i_j$ with delays on $\pi^i_{j'}$. Observe that outside the segment $\pi^i_{s+1}\ldots \pi^i_{t}$, the plans $P'$ and $P''$ are identical, and in particular have the same budget and are both non-colliding. 

By repeating this for all vertices that are not in ICG, we end up with a non-colliding plan in ICG, as desired.
\end{proof}



\end{document}